\pdfoutput=1

\documentclass[11pt]{article}

\usepackage[final]{acl}

\usepackage{times}
\usepackage{latexsym}

\usepackage[T1]{fontenc}

\usepackage[utf8]{inputenc}

\usepackage{microtype}

\usepackage{inconsolata}

\usepackage{graphicx}
\usepackage{enumitem}
\usepackage{tablefootnote}
\usepackage{amsfonts}
\usepackage{amsmath}
\usepackage{amssymb}
\usepackage{multirow}
\usepackage[linesnumbered,ruled,lined]{algorithm2e}
\usepackage{hyperref}

\newcommand{\mat}[1]{{\bf #1}}   
\newcommand{\method}{\textsc{MatMcd}}
\newcommand{\modulea}{\textsc{DA-agent}}
\newcommand{\moduleb}{\textsc{CC-agent}}

\SetCommentSty{mycommfont}

\newcommand\blfootnote[1]{%
  \begingroup
  \renewcommand\thefootnote{}\footnote{#1}%
  \addtocounter{footnote}{-1}%
  \endgroup
}

\usepackage{booktabs}

\newcommand{\zc}[1]{\textcolor{orange}{#1}}
\newcommand{\nop}[1]{}
\newcommand{\var}[1]{\textcolor{blue}{#1}}

\usepackage{listings}
\usepackage{todonotes}

\title{Exploring Multi-Modal Data with Tool-Augmented LLM Agents\\for Precise Causal Discovery}

\author{ChengAo Shen\textsuperscript{\rm 1}, Zhengzhang Chen\textsuperscript{\rm 2}\textsuperscript{*}, Dongsheng Luo\textsuperscript{\rm 3}, Dongkuan Xu\textsuperscript{\rm 4},\\{\bf Haifeng Chen\textsuperscript{\rm 2},} {\bf Jingchao Ni\textsuperscript{\rm 1}\textsuperscript{*}}\\
\textsuperscript{\rm 1}University of Houston, \textsuperscript{\rm 2}NEC Laboratories America,\\
\textsuperscript{\rm 3}Florida International University, \textsuperscript{\rm 4}North Carolina State University\\
\textsuperscript{\rm 1}\texttt{\{cshen9, jni7\}@uh.edu}, \textsuperscript{\rm 2}\texttt{\{zchen, haifeng\}@nec-labs.com},\\
\textsuperscript{\rm 3}\texttt{dluo@fiu.edu}, \textsuperscript{\rm 4}\texttt{dxu27@ncsu.edu}}

\begin{document}
\maketitle

\begin{abstract}
Causal 
discovery\blfootnote{{\bf\textsuperscript{*}}Corresponding authors.} is an imperative foundation for decision-making across domains, such as smart health, AI for drug discovery and AIOps. Traditional statistical causal discovery methods, while well-established, predominantly rely on observational data and often overlook the semantic cues inherent in cause-and-effect relationships. The advent of Large Language Models (LLMs) has ushered in an affordable way of leveraging the semantic cues for knowledge-driven causal discovery, but the development of LLMs for causal discovery lags behind other areas, particularly in the exploration of 
multi-modal data. To bridge the gap, we introduce \method, a multi-agent system powered by tool-augmented LLMs. \method\ has two key agents: a Data Augmentation agent that retrieves and processes modality-augmented data, and a Causal Constraint agent that integrates multi-modal data for knowledge-driven 
reasoning. 
The proposed design of the inner-workings ensures successful cooperation of the agents. Our empirical study across seven datasets suggests the significant potential of multi-modality enhanced causal discovery.\footnote{The code of \method\ is available at \url{https://github.com/D2I-Group/matmcd}.}

\end{abstract}

\section{Introduction}

\begin{figure}[!t]
\centering
\includegraphics[width=\columnwidth]{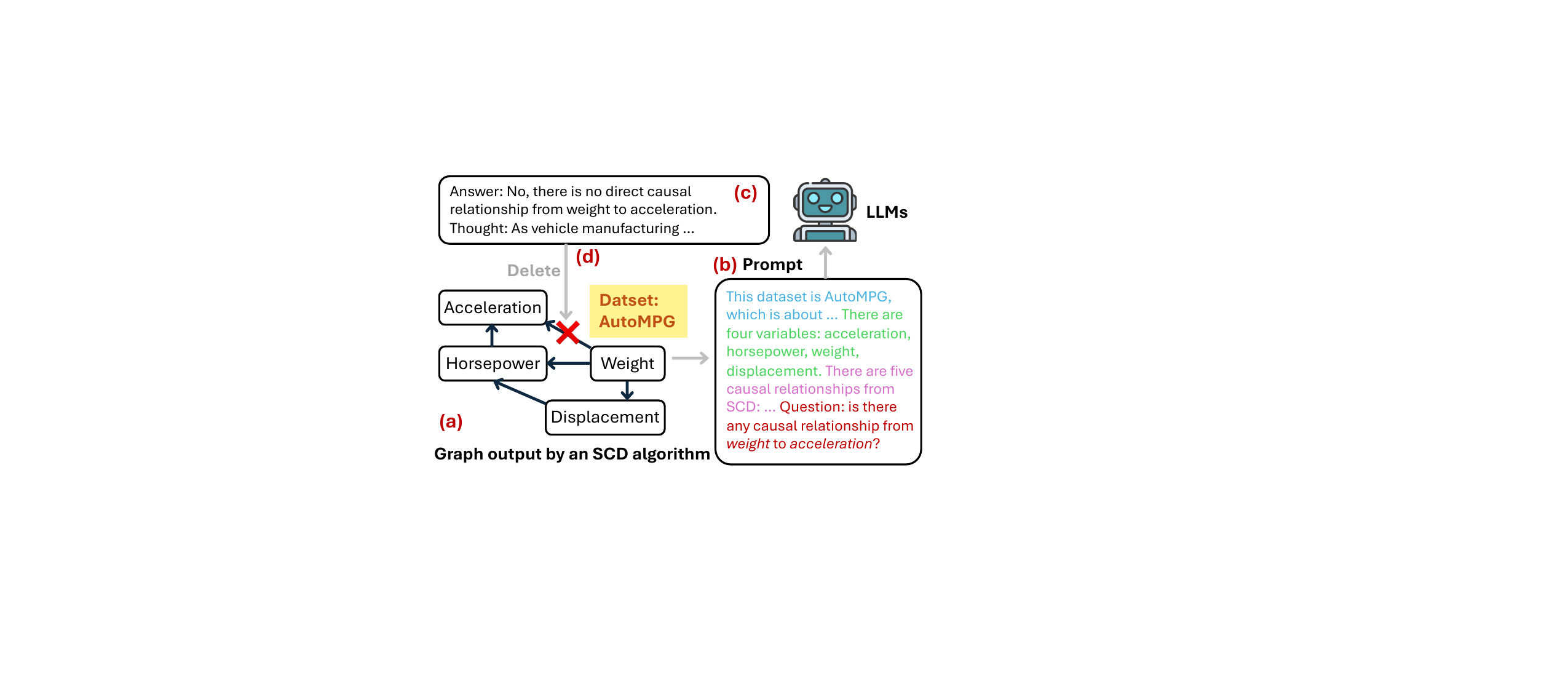}
\caption{An illustration of the general process of LLM based causal discovery: (a) a causal graph is estimated by an SCD algorithm; (b) a prompt is generated {\em w.r.t.} the graph; (c) LLMs reason about the causal structures; (d) the graph is refined based on the LLM response.
}\label{fig.introduction}
\vspace{-0.3cm}
\end{figure}

Identifying cause-and-effect relationships in complex systems is crucial for a variety of applications, including neuralgia diagnosis in medicine~\cite{tu2019neuropathic}, protein pathway analysis in computational biology~\cite{sachs2005causal}, and root cause locating in microservice architectures~\cite{wang2023incremental}. These causal insights significantly benefit emerging fields such as smart health, AI-driven drug discovery, and AIOps. The process of discovering such relationships from observational data, known as {\em causal discovery} \cite{yu2025causaleval}, typically generates a Directed Acyclic Graph (DAG). In this graph, edges represent the existence and direction of causal relationships between variables, as illustrated in Fig. \ref{fig.introduction}(a). This DAG not only governs the data generation process but also enhances the understanding of inter-variable influences, serving as the foundation for many downstream decision-making tasks~\cite{nguyen2023causal}. As such, constructing accurate causal graphs is essential to the reliability of subsequent analyses.

\nop{Discovering cause-and-effect relationship between variables underlying complex systems 
is significant to a variety of applications, including the diagnosis of 
neuralgia in medicine~\cite{tu2019neuropathic}, pathway analysis of proteins in 
biology~\cite{sachs2005causal}, and root cause analysis in microservice systems~\cite{zheng2024lemma}. It is substantially beneficial for emerging domains such as smart health, AI for drug discovery, and AIOps. This process, namely {\em causal discovery}, usually produces a Directed Acyclic Graph (DAG), with edges representing the existence and direction of causal relationships between variables, as illustrated in Fig. \ref{fig.introduction}(a). This graph governs the generation of data, enables understanding of the mutual influences of variables, and serves as the foundation for many downstream decision-making tasks \cite{nguyen2023causal}. 
As such, ensuring correct causal graphs is essential for the reliability of subsequent analysis.}

Conventional methods 
primarily rely on data-driven statistical causal discovery (SCD), 
which can be categorized as non-parametric \cite{spirtes1991algorithm,silander2006simple,huang2018generalized,xie2020generalized} 
and semi-parametric~\cite{shimizu2006linear,shimizu2011directlingam,zheng2018dags,tu2022optimal}. These methods estimate causal relationships by analyzing the observational data of variables, but they overlook the semantic and contextual cues of the variables, resulting in suboptimal outcomes~\cite{takayama2024integrating}.

\nop{Conventional methods for causal discovery include data-driven statistical causal discovery (SCD) algorithms that are either non-parametric 
\cite{spirtes1991algorithm,silander2006simple,huang2018generalized,xie2020generalized} or semi-parametric 
\cite{shimizu2006linear,shimizu2011directlingam,zheng2018dags,tu2022optimal}. These methods use data samples emitted by variables to estimate the causal relationships among them, but neglect the semantic meanings and contextual cues of variables, leading to suboptimal solutions \cite{takayama2024integrating}.}

Commonsense and domain knowledge are invaluable for identifying cause-and-effect relationships among semantically meaningful variables. In light of this, growing research attention has been drawn to elicit such information for causal discovery. Large Language Models (LLMs), as praised by its astonishing reasoning ability drawing on extensive knowledge 
acquired from large-scale training \cite{brown2020languagemodelsfewshotlearners,achiam2023gpt}, now become a promising and cost-effective 
source of expert knowledge to aid causal discovery. For example, by merely prompting with variable names and some contextual cues, LLMs have been shown to infer meaningful causal relationships 
\cite{ban2023causal,jiralerspong2024efficient}. More recently, hybrid approaches were introduced to combine LLMs with data-driven SCD algorithms, achieving enhanced accuracy in causal discovery \cite{takayama2024integrating,khatibi2024alcm}. However, most existing methods have yet to fully harness the potential of modern LLMs, particularly the agent systems built upon tool-augmented LLMs.

\nop{Because commonsense and domain knowledge are invaluable sources in identifying the cause-and-effect between semantically meaningful variables, a lot of recent research attention has been drawn to leverage such information for causal discovery. Recently, Large Language Models (LLMs) have been found with astonishing reasoning ability based on extensive knowledge 
acquired from large-scale training \cite{brown2020languagemodelsfewshotlearners,achiam2023gpt}, thus presenting a promising 
way for obtaining 
affordable expert knowledge for causal discovery. For example, by merely prompting with the variable names and some contextual cues, LLMs were found being able to infer meaningful causal relationships \cite{vashishtha2023causal,ban2023causal,jiralerspong2024efficient}. The more recent efforts introduced hybrid approaches combining LLMs and data-driven SCD algorithms for enhanced accuracy of causal discovery \cite{takayama2024integrating,khatibi2024alcm}. 
However, most of the existing methods haven't explored the full capability of modern LLMs, particularly the agent systems built upon tool-augmented LLMs.}

An LLM agent is typically equipped with memory, reasoning, planning, and access to external tools such as calculators, search engines, and code compilers, rendering it superior in problem-solving compared to vanilla LLMs \cite{yao2022react}. As a single-agent system may be hallucination-prone even with self-reflection 
\cite{li2023camel,shinn2024reflexion}, 
multi-agent systems were introduced, 
which rivals more advanced models by combining multiple weaker agents. Despite the fast progress, LLM agents for causal discovery remains underexplored. The most relevant effort to date introduced a multi-agent system with debating LLMs for synergistic causal 
discovery \cite{le2024multi}. However, this approach has never explored the potential of {\em multi-modality} in data -- a feature that agent systems are well-equipped to handle.

\nop{An LLM agent can be seen as an entity with memory, reasoning, planning, and 
external tools such as a calculator, search engine, and code compiler, which has demonstrated superior 
problem-solving performance 
over vanilla LLMs \cite{yao2022react}. 
Moreover, a multi-agent system could excel a single-agent system as the latter may suffer from hallucinations even with self-reflection \cite{li2023camel,shinn2024reflexion,madaan2024self}. Despite the encouraging progress, 
LLM agents for causal discovery are almost untapped. The most relevant is a work that introduced a multi-agent system 
with several debating LLMs for synergistic causal inference 
\cite{le2024multi}, 
but it has never explored the {\em multi-modality} of data that an agent system can easily take advantage of.}

As  Fig. \ref{fig.introduction} shows, hybrid methods for causal discovery typically prompt LLMs with the prior causal graph produced by some SCD algorithm, appended by some meta-data ({\em e.g.}, variable names, dataset titles) as contexts. However, these inputs may fall short in fully activating the reasoning ability of LLMs. Inspired by the observation that 
abundant semantic data from external sources, such as webs and logs, can serve as an {\em additional modality to the observational data (e.g., time series)} 
for improving prompts, we propose a \underline{M}ulti-\underline{A}gent system with \underline{T}ool-augmented LLMs for exploring \underline{M}ulti-modal 
data enhanced \underline{C}ausal \underline{D}iscovery (\method).

\nop{As in Fig. \ref{fig.introduction}, the existing hybrid methods for causal discovery typically prompt LLMs with the prior causal graph produced by an SCD algorithm, with some meta data as the contextual cues such as variable names and dataset titles, which, however, may be limited in stimulating LLMs' reasoning ability. We observed the abundant semantic data retrieved from external sources, such as webs and logs, could constitute an {\em additional data modality to the causal graph} and largely enrich the prompt instructions. Therefore, in this work, 
we propose a \underline{M}ulti-\underline{A}gent system with \underline{T}ool-augmented LLMs for exploring \underline{M}ulti-modal enhancement of \underline{C}ausal \underline{D}iscovery (\method).}

Specifically, \method\ is designed as a framework for refining causal graphs generated by SCD algorithms, involving two key agents: (1) a Data Augmentation Agent (\modulea, $\S$\ref{sec.data_agent}), and (2) a Causal Constraint Agent (\moduleb, $\S$\ref{sec.cc_agent}). Given a causal graph output by some SCD algorithm, \modulea\ integrates meta-data ({\em i.e.}, variable names and dataset titles) and calls tools such as web search APIs or offline log lookup APIs for iterative, reflection-based retrieval of augmented (textual) data, which is summarized as a compact cue in a different modality from the graphs. Upon receiving the augmented data, \moduleb\ combines it with the topological structure of the prior causal graph to infer the causal relationships among variables. Both agents comprise multiple cooperative LLMs with 
well-crafted mechanisms to handle sub-tasks such as tool-calling, memorizing, reasoning, and summarization. Retrieval Augmented Generation (RAG) \cite{lewis2020retrieval} components are used where efficient memory is essential. In our experiments, we compared \method\ with state-of-the-art (SOTA) baseline methods across five benchmark datasets and two public AIOps datasets of microservice systems. The results demonstrate substantial improvements in causal discovery by incorporating multi-modal data. 
The main contributions of this work are summarized as follows:
\begin{itemize}[noitemsep,topsep=0pt,leftmargin=*]
\item We propose to explore the problem of 
multi-modal data enhanced causal discovery via LLM agents, which is significant yet less studied.
\item We introduce \method, a novel framework of multi-agent, as a testbed for assessing the effectiveness of 
multi-modal data in causal discovery.
\item We perform extensive experiments on a variety of datasets, 
where \method\ reduces causal 
discovery errors (NHD) by up to 66.7\% and improves root cause locating (MAP@10) by up to 83.3\% over the best baselines, suggesting the potential of 
multi-modal data in causal discovery.

\end{itemize}

\section{Related Work}
\noindent{\bf Causal Discovery Methods} are mostly conventional data-driven SCD algorithms, including non-parametric methods 
\cite{spirtes1991algorithm,chickering2002optimal,silander2006simple,huang2018generalized,xie2020generalized}
and semi-parametric methods \cite{shimizu2006linear,hoyer2008nonlinear,shimizu2011directlingam,zheng2018dags,rolland2022score,tu2022optimal}. These methods rely on observational data as input but cannot leverage the semantics of variables. Recently, knowledge-driven methods have been found promising for causal discovery. Some early efforts use LLMs by simply prompting variable names and dataset titles \cite{kiciman2023causal,zevcevic2023causal,chen2024causal,jiralerspong2024efficient}. Then hybrid approaches that integrate SCD algorithms with LLMs were introduced \cite{ban2023causal,vashishtha2023causal,khatibi2024alcm,takayama2024integrating} and found to be more effective than pure LLM-based methods. More recently, a multi-agent system-based approach was proposed \cite{le2024multi} to explore the impacts of debating LLMs. However, none of these methods has explored the potential of multi-modal data in the LLM-based causal discovery process.

\nop{\noindent{\bf Causal Discovery Methods}. Conventional methods are mostly data-driven SCD algorithms including both non-parametric methods \cite{spirtes1991algorithm,chickering2002optimal,silander2006simple,yuan2013learning,huang2018generalized,xie2020generalized} and semi-parametric methods \cite{shimizu2006linear,hoyer2008nonlinear,shimizu2011directlingam,zheng2018dags,rolland2022score,tu2022optimal}. These methods rely on metrics data as input but cannot leverage the semantics of variables. Recently, knowledge-driven approaches using LLMs have been found promising for causal discovery. Some early efforts to use LLMs have been made by simply prompting variable names and data set titles~\cite{kiciman2023causal,zevcevic2023causal,chen2024causal,jiralerspong2024efficient}. Then hybrid approaches that integrate SCD algorithms with LLMs were introduced \cite{ban2023causal,vashishtha2023causal,khatibi2024alcm,takayama2024integrating} and were found to be more effective than pure LLM-based methods. More recently, a multi-agent system-based approach was proposed \cite{le2024multi} to explore the impacts of debating LLMs. However, none of these methods has explored the potential of multi-modal data in the LLM based causal discovery process.}

\vspace{0.1cm}

\nop{\noindent{\bf LLM Agents.} An LLM agent framework is typically composed of user request, agent, plan, memory, and tools, with the agent lies in the core. Planning can leverage techniques such as Chain-of-Thought (CoT) \cite{wei2022chain}, ReAct \cite{yao2022react}, and Reflexion \cite{shinn2024reflexion}. Tools enable interaction with the environment, and frameworks like MRKL \cite{karpas2022mrkl}, Toolformer \cite{schick2024toolformer}, Function Calling \cite{function_calling_2024}, and HuggingGPT \cite{shen2024hugginggpt} exemplify paradigms that integrate tools for problem-solving. For complex tasks, multi-agent systems often outperform single-agent systems. The three primary categories of multi-agent systems include cooperative agents \cite{qian2023communicative,chen2024scalable}, competitive agents \cite{zhao2023competeai}, and debating agents \cite{li2023camel,liang2023encouraging,xiong2023examining}. In this work, we focus on a cooperative multi-agent system, where agents are coordinated and exchange information to collaboratively enhance solutions toward a shared goal.}

\noindent{\bf LLM Agents} are typically equipped with plan, memory and tools. 
Planning can use techniques such as CoT \cite{wei2022chain}, ReAct \cite{yao2022react}, and Reflexion \cite{shinn2024reflexion}. Tools endow agents with the ability to interact with environments. 
MRKL \cite{karpas2022mrkl}, Toolformer \cite{schick2024toolformer}, Function Calling \cite{function_calling_2024}, and HuggingGPT \cite{shen2024hugginggpt} exemplify paradigms that integrate tools for problem-solving. For complex tasks, multi-agent systems are promising. The primary categories of multi-agent systems include cooperative agents \cite{qian2023communicative,chen2024scalable}, competitive agents \cite{zhao2023competeai} and debating agents \cite{li2023camel,liang2023encouraging,xiong2023examining}. In this work, we investigate a cooperative multi-agent system where the agents are coordinated to 
enhance a solution towards a shared goal.

\section{Methodology}\label{sec.method}

\begin{figure*}[!t]
\centering
\includegraphics[width=\linewidth]{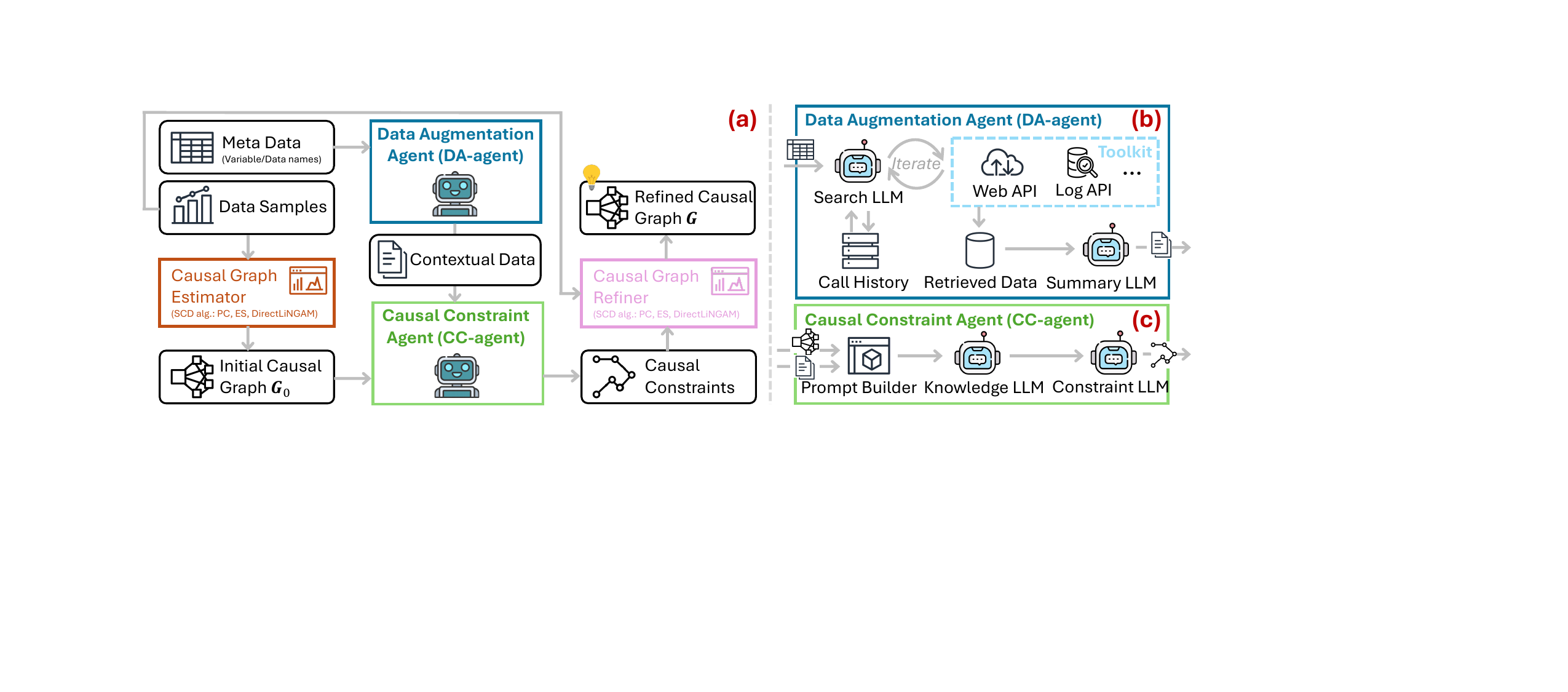}
\caption{An illustration of the \method\ framework: (a) an overview of the framework, (b) the inner working of \modulea, and (c) the inner working of \moduleb.}\label{fig.method}
\end{figure*}

Fig. \ref{fig.method} is an overview of the proposed \method\ system, which has four key components: (1) Causal Graph Estimator ($\S$\ref{sec.cge}); (2) Data Augmentation Agent (\modulea, $\S$\ref{sec.data_agent}); (3) Causal Constraint Agent (\moduleb, $\S$\ref{sec.cc_agent}); and (4) Causal Graph Refiner ($\S$\ref{sec.cgr}). Next, we will first introduce some notations and then elaborate on each of the components in an order subject to the flow of data.

\vspace{0.1cm}

\nop{\noindent{\bf Notations.} Let $\mathcal{V} = \{v_{1}, \dots, v_{n}\}$ represent a set of $n$ variables ({\em e.g.}, 4 variables in Fig. \ref{fig.introduction}(a)). Each variable $v_{i} \in \mathcal{V}$ is associated with a set of observed data samples $\mat{v}_{i} = \{v_{i1}, \dots, v_{im}\}$, where $m$ is the number of samples. Depending on the application, the observational data could either be random samples or a length-$m$ time series generated by each variable. In addition to the observational data, meta-data is often available. In this work, we assume the availability of a minimal set of meta-data $\mathcal{D} = \{\mathsf{s}, \mathcal{Z}\}$, where $\mathsf{s}$ is a descriptive title of the dataset ({\em e.g.}, ``AutoMPG'' in Fig. \ref{fig.introduction}(a)), and $\mathcal{Z} = \{\mathsf{z}_{1}, \dots, \mathsf{z}_{n}\}$ is a set of descriptive names for the variables in $\mathcal{V}$, where $\mathsf{z}_{i}$ corresponds to variable $v_{i}$ ({\em e.g.}, ``Acceleration'' in Fig. \ref{fig.introduction}(a)).}

\noindent{\bf Notations.} Suppose there is a set of $n$ variables $\mathcal{V}=\{v_{1}$, ..., $v_{n}\}$ ({\em e.g.}, 4 variables in Fig. \ref{fig.introduction}(a)), each variable $v_{i}$ is associated with a set of observed data samples $\mat{v}_{i}=\{v_{i1}, ..., v_{im}\}$ where $m$ is the number of samples. The observational data could be random samples or a length-$m$ time series emitted by each variable, depending on the application. In many cases, meta-data is available. In this work, we assume a minimal set of meta-data $\mathcal{D}=\{\mathsf{s}, \mathcal{Z}\}$ where $\mathsf{s}$ is a descriptive title of the dataset ({\em e.g.}, ``AutoMPG'' in Fig. \ref{fig.introduction}(a)) and $\mathcal{Z}=\{\mathsf{z}_{1}, ..., \mathsf{z}_{n}\}$ includes the descriptive name $\mathsf{z}_{i}$ of each variable $v_{i}$ in $\mathcal{V}$ ({\em e.g.}, ``Acceleration'' in Fig. \ref{fig.introduction}(a)).

\vspace{0.1cm}

\nop{\noindent{\bf The Task.} Given a set of variables $\mathcal{V}$ and a meta-data $\mathcal{D}$, the objective is to construct a DAG $\mat{G}=(\mathcal{V}, \mathcal{E})$, where each variable $v_{i} \in \mathcal{V}$ is represented as a node, and $\mathcal{E} \subseteq \mathcal{V} \times \mathcal{V}$ is the set of {\em directed} edges. A directed edge $(v_{i}, v_{j}) \in \mathcal{E}$ signifies a causal relationship from $v_{i}$ to $v_{j}$. The primary goal of this work is to accurately infer the causal relationships encoded in $\mathcal{E}$ among the $n$ variables in $\mathcal{V}$. \zc{We should cover LLM or leveraging LLM in the problem statement.}}

\noindent{\bf The Task.} Based on $\{\mat{v}_{1}, ..., \mat{v}_{n}\}$ and $\mathcal{D}$, we want to construct a DAG, $\mat{G}=(\mathcal{V}, \mathcal{E})$, where each variable $v_{i}$ in $\mathcal{V}$ is a node in the graph, and $\mathcal{E}\subseteq\mathcal{V}\times\mathcal{V}$ is the set of {\em directed} edges with $(v_{i}, v_{j}) \in \mathcal{E}$ signifying a causal relationship from $v_{i}$ to $v_{j}$. The goal of this work is to infer accurate causal relationships in $\mathcal{E}$ among the set of $n$ variables in $\mathcal{V}$.

\subsection{Causal Graph Estimator}\label{sec.cge}

\nop{Similar to hybrid approaches~\cite{takayama2024integrating,khatibi2024alcm}, the Causal Graph Estimator serves as an initializer for the causal graph and is built upon data-driven statistical causal discovery (SCD) algorithms. Its primary goal is to estimate an initial causal graph, $\mat{G}_{0} = (\mathcal{V}, \mathcal{E}_{0})$, purely from observational data, $\{\mat{v}_{1}, \mat{v}_{2}, \ldots, \mat{v}_{n}\}$, without relying on any additional information. Notably, $\mathcal{V}$ remains unchanged throughout the proposed framework, as our focus lies on refining the causal relationships in $\mathcal{E}$ to enable more accurate causal discovery.}

Similar to the hybrid approaches \cite{takayama2024integrating,khatibi2024alcm}, Causal Graph Estimator serves as an initializer of causal graph and is built upon data-driven SCD algorithms, with the aim of estimating an initial causal graph $\mat{G}_{0}=(\mathcal{V}, \mathcal{E}_{0})$ purely from the observational data $\{\mat{v}_{1}$, ..., $\mat{v}_{n}\}$ without accessing any other information. Here $\mathcal{V}$ is not subscript as it will be kept intact throughout the proposed framework and our focus is on the alteration of the causal relationships in $\mathcal{E}$ for accurate causal discovery.

\nop{Our framework is flexible in its choice of SCD algorithms. In this work, we explore the feasibility of employing three widely used algorithms, each representing a distinct category: (1) Constraint-based method -- the Peter-Clark (PC) algorithm~\cite{spirtes1991algorithm}, which is non-parametric; (2) Score-based method -- the Exact Search (ES) algorithm~\cite{yuan2013learning}, which is also non-parametric; and (3) Constrained functional causal models -- DirectLiNGAM~\cite{shimizu2011directlingam}, which is semi-parametric. We leave the exploration of other SCD algorithms to future work, as it is beyond the scope of this work. Inspired by recent LLM prompting techniques for graphs \cite{wang2024can}, the output edges $\mathcal{E}_{0}$ from the SCD algorithm are embedded as an adjacency list within the prompt generated by the prompt builder module of the \moduleb\ (see $\S$\ref{sec.cc_agent} in Fig.~\ref{fig.method}). This prompt also integrates semantics-rich data modalities retrieved by the \modulea\ (see $\S$\ref{sec.data_agent}).}

Our framework is flexible to the choice of SCD algorithms. In this work, we investigate the feasibility of employing three widely used algorithms, each of which is a representative of a category: (1) Constraint-based method -- Peter-Clark (PC) algorithm \cite{spirtes1991algorithm} which is non-parametric; 
(2) Score-based method -- Exact Search (ES) algorithm \cite{yuan2013learning} which is non-parametric; and (3) Constrained functional causal models -- DirectLiNGAM \cite{shimizu2011directlingam} which is semi-parametric. 
We leave the exploration of other SCD algorithms in our future work as it is out of the scope of this work. Inspired by the recent LLM prompting techniques for graphs \cite{wang2024can}, the output edges $\mathcal{E}_{0}$ of the SCD algorithm will be embedded as an adjacency list in the prompt generated by the prompt builder module of the \moduleb\ ($\S$\ref{sec.cc_agent}) in Fig. \ref{fig.method}, which will also integrate the semantics-rich data modality retrieved by the \modulea\ ($\S$\ref{sec.data_agent}).

\subsection{Data Augmentation Agent (\modulea)}\label{sec.data_agent}
The goal of \modulea\ is to retrieve semantics-rich contextual data pertinent to the initial causal graph $\mat{G}_{0}$, such as web documents and log files about the variables, as an additional modality to observational data for prompting 
\moduleb\ 
($\S$\ref{sec.cc_agent}). As illustrated in Fig.~\ref{fig.method}(b), \modulea\ comprises a Search LLM and a Summary LLM.

\nop{The goal of \modulea\ is to retrieve semantics-rich contextual data about the initial causal graph $\mat{G}_{0}$, which could be web docs and log files about the variables, 
as an additional modality for prompting the \moduleb\ ($\S$\ref{sec.cc_agent}). As illustrated by Fig. \ref{fig.method}(b), \modulea\ consists of a Search LLM and a Summary LLM.}

\vspace{0.1cm}

\nop{\noindent{\bf Search LLM.} The Search LLM is equipped with a set of tools for data retrieval. In this work, we focus on two key tools: a web search API, which serves as a general-purpose tool for retrieving contextual data from external sources, and a log lookup API, designed for applications where a domain-specific database is available, such as process logs used in root cause analysis for microservice systems in AIOps \cite{zheng2024lemma}. The \modulea\ is flexible and extendable, allowing integration of additional application-specific tools, such as a Wikipedia API or a code lookup API, which we leave for future exploration.}

\noindent{\bf Search LLM.} The Search LLM has access to a set of tools for data search. In this work, we focus on a web search API as a general tool for retrieving contextual data from external sources, and a log lookup API for applications where a domain-specific database is available such as the process logs in root cause analysis for microservice systems in AIOps \cite{zheng2024lemma}. The \modulea\ is flexible to the toolkit and is extensible for a wide scenarios by including other application-specific tools such as Wikipedia API and code lookup API, which we leave for future exploration.

As shown in Fig.~\ref{fig.method}, upon receiving the meta-data $\mathcal{D} = \{\mathsf{s}, \mathcal{Z}\}$ about the causal graph, the Search LLM first checks its {\em calling history memory} to decide whether to initiate a new tool call. If a new call is needed, the Search LLM invokes a search tool API to retrieve additional data using a prompt that includes the dataset title $\mathsf{s}$ and variable names $\mathsf{z}_{1}$, ..., $\mathsf{z}_{n}$. This search action is then recorded in the memory for future reference. 
Since our focus is on search tools and the search action involves generating a query, 
the generated query is added to the memory. In subsequent iterations, all previously recorded queries are examined to prevent redundant queries. This process continues iteratively until the Search LLM determines that no further tool calls are necessary, terminating the loop. 

To enable this iterative search, the prompt is designed based on self-reflection techniques \cite{shinn2024reflexion,madaan2024self}, where the LLM assesses whether additional queries are needed based on the comprehensiveness of the historical queries. The loop terminates when the LLM concludes with a ``No query needed'' response. Compared to single-round searches, this iterative process proves crucial for retrieving relevant and comprehensive data, especially in domains where variable-specific information is challenging to locate ({\em e.g.}, medicine). However, for lookup APIs, iteration is unnecessary. Thus, the retrieved data through these tools constitutes an additional textual modality for the causal graph. The prompt used for the Search LLM is provided in Appendix~\ref{app:DA_agent prompt}.

\begin{figure}[!t]
\centering
\includegraphics[width=\columnwidth]{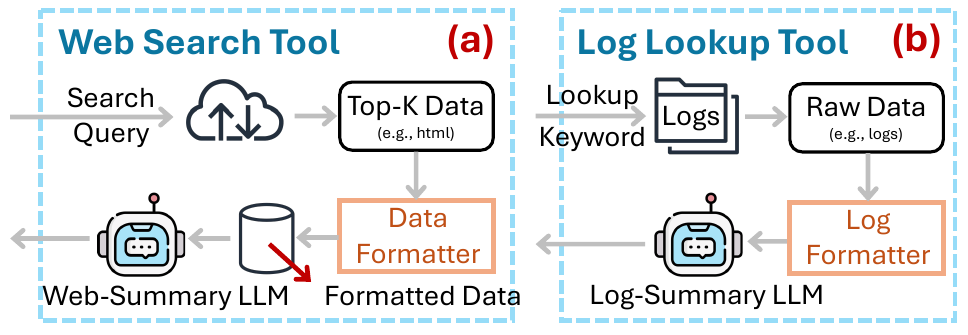}
\caption{The search tool preparation in \modulea\ for (a) web search tool, and (b) log lookup tool.}\label{fig.tool}
\vspace{-0.4cm}
\end{figure}

\vspace{0.1cm}

\nop{\noindent{\bf Tool Preparation.} Fig.~\ref{fig.tool} illustrates the preparation of two tools: the Web Search tool and the Log Lookup tool. For the Web Search tool, we utilize the Google Search API, where the queries are generated by the Search LLM as described earlier. The top retrieved webpages are processed by a data formatter to remove formatting artifacts (\textit{e.g.}, HTML tags), resulting in plain text documents that are stored in memory. These documents are then summarized into concise descriptions using a {\em Web-Summary LLM}. In contrast, the Log Lookup tool retrieves logs directly based on a variable name used as a keyword, bypassing the need for memory storage. The retrieved logs, which may include unnecessary formatting (\textit{e.g.}, log templates) and can be lengthy, are processed by removing extraneous elements and summarized by a {\em Log-Summary LLM}.}

\noindent{\bf Tool Preparation.} \label{sec_tool_preparation} Fig. \ref{fig.tool} illustrates our preparation of the Web Search tool and the Log Lookup tool. In the former, we employ Google search API where the query is generated by the Search LLM as aforementioned. The retrieved top webpages will be de-formatted ({\em e.g.}, removing HTML tags) by a data formatter and the resultant plain docs will be stored in a memory. Then a {\em Web-Summary LLM} is employed to summarize the docs into a concise description. In contrast, the Log Lookup tool uses exact lookup, {\em i.e.}, with a variable name as the keyword, its corresponding log can be retrieved directly. Thus the memory can be removed and the retrieved log, which still needs de-formatting ({\em e.g.}, removing log templates) and could be lengthy, will be summarized by a {\em Log-Summary LLM}.

\vspace{0.1cm}

\noindent{\bf Summary LLM.} The data retrieved by the Search LLM is iteratively added to the {\em Retrieved Data Memory}, as shown in Fig. \ref{fig.method}(b). Upon loop termination, a {\em Summary LLM} summarizes the retrieved data into three types of cues: (1) description of the dataset; (2) description of each variable in the graph; and (3) relationships between the variables. Since the size of the retrieved data from iterative searches may exceed the LLM's context window, we adopt an efficient summarization approach using RAG \cite{lewis2020retrieval}. The retrieved data is divided into indexed document chunks, implemented with \texttt{LlamaIndex} \cite{LiuLlamaIndex2022} using \texttt{text-embedding-ada-002} for chunk indexing and Maximum Inner Product Search (MIPS) for retrieving relevant chunks. An example summary is provided in Appendix \ref{app:DA-Agent example}. The resulting summary serves as 
{\em contextual data} for the initial causal graph $\mat{G}_{0}$ to prompt \moduleb.

\nop{\noindent{\bf Summary LLM.} The retrieved data by Search LLM will be iteratively added to the {\em Retrieved Data memory} as illustrated in Fig. \ref{fig.method}(b). Upon the termination of the loop, a {\em Summary LLM} will be asked to summarize the retrieved data into three types of cues: (1) description of the dataset; (2) description of each variable in the graph; and (3) relationships between the variables. Because the amount of the retrieved data could be too large to fit an LLM's context window due to the iterative search, for efficient summarization, we structured the memory by RAG \cite{lewis2020retrieval}, where the retrieved data were split into document chunks with indices. We implemented the RAG by \texttt{LlamaIndex} \cite{LiuLlamaIndex2022} with \texttt{text-embedding-ada-002} for indexing the chunks and Maximum Inner Product Search (MIPS) for searching the relevant chunks for summarization. 
An example summary can be found in Appendix \ref{app:DA-Agent example}. The output summary will used as a {\em contextual data modality} to the initial causal graph $\mat{G}_{0}$ for prompting \moduleb.}

\vspace{0.1cm}

\noindent{\bf Remark.} Since there is a risk of retrieving web content that may leak ground truth causal graphs for certain datasets, we conducted thorough screening in our RAG implementation of the {\em Retrieved Data Memory} to prevent such information leaks.

\nop{\noindent{\bf Remark.} Since there is a risk of retrieving webs that may leak the ground truth causal graphs in some datasets, in our RAG implementation of the {\em Retrieved Data memory}, we conducted thorough screening to avoid any such information leaks.}

\subsection{Causal Constraint Agent (\moduleb)}\label{sec.cc_agent}

In addition to the external knowledge, our method leverages the factual knowledge stored in LLMs, acquired during pre-training. To achieve this, \moduleb\ is designed based on the Two-Stage Prompting framework of zero-shot Chain-of-Thought (ZSCOT)~\cite{kojima2022large}. First, a prompt builder integrates $\mat{G}_{0}$, represented as an adjacency list, with contextual data from \modulea\ to prompt a {\em Knowledge LLM}. The Knowledge LLM is tasked with explaining each (non-)existing causal relationship in the initial causal graph $\mat{G}_{0}$ based on the contextual data and its own knowledge. These explanations, which could either support or refute the causal relationships, are used to prompt a {\em Constraint LLM} in the second stage to draw a conclusion on the existence of each relationship ({\em i.e.}, ``Yes''/``No''). To address potential uncertainty in the conclusions, we adopt the Top-K-Guess technique \cite{tian2023just} to elicit verbal confidence. This approach, found to be more reliable than sampling-based likelihood estimation \cite{xiong2024can}, quantitatively evaluates the likelihood of each causal relationship. Among the Top-K guesses, the most confident one is selected as the final conclusion for each causal relationship.

\nop{In addition to the external knowledge, our method also leverages the factual knowledge of LLMs that were acquired by pre-training and stored in their parameters. To this end, \moduleb\ was designed with the Two-Stage Prompting idea of zero-shot Chain-of-Thought (ZSCOT) \cite{kojima2022large}. First, a prompt builder integrates $\mat{G}_{0}$ as an adjacency list with the contextual data (from the \modulea) for prompting a {\em Knowledge LLM}. The Knowledge LLM is asked for explaining each of the (non-)existing causal relationships in the initial causal graph $\mat{G}_{0}$ based on the contextual data and its own knowledge. The explanation could either support or deny the causal relationship and will be used for (the 2nd-stage) prompting a {\em Constraint LLM} for drawing a conclusion on its existence ({\em i.e.}, ``Yes''/``No''). Additionally, to account for the potential uncertainty of the conclusion, we adopted the Top-K-Guess technique \cite{tian2023just} for eliciting verbal confidence, which was found more reliable than sampling-based likelihood estimation \cite{xiong2024can}, for quantitatively evaluating the likelihood of a causal relationship. Among the Top-K guesses, the most confident one will be used as the conclusion on each causal relationship.}

\subsection{Causal Graph Refiner}\label{sec.cgr}

To ensure the final causal graph $\mat{G}$ is acyclic, the edge set $\mathcal{E}_{0}$ is not directly modified based on the (non-)existence constraints from \moduleb. Instead, the SCD algorithm used in the Causal Graph Estimator ($\S$\ref{sec.cge}) is rerun with these constraints imposed to generate a new causal graph $\mat{G}$. Specifically, upon receiving the (non-)existence constraints, a {\em constraint matrix} $\mat{C}\in \mathbb{R}^{n\times n}$ is constructed, where $\mat{C}_{ij}=1$ if \moduleb\ indicates a causal effect from $v_{i}$ to $v_{j}$, and $\mat{C}_{ij}=0$ otherwise. 

The most representative SCD algorithms, including PC, ES, and DirectLiNGAM, are designed to incorporate such a constraint matrix $\mat{C}$ as input alongside observational data $\{\mat{v}_{1}, \dots, \mat{v}_{n}\}$. This ensures that the generated causal graph complies with the constraints in $\mat{C}$ to varying extents, enabling the production of a refined causal graph $\mat{G}$ that is both consistent with the imposed constraints and a directed acyclic graph (DAG).

\vspace{0.2cm}

\noindent{\bf Remark.} Compared to the existing approaches that utilize LLMs for causal discovery~\cite{takayama2024integrating,khatibi2024alcm}, the key novelty of the proposed \method\ lies in its exploration of \modulea\ for multi-modal 
data enhanced causal discovery. An algorithmic summary of the proposed workflow is provided in Appendix~\ref{app:alg}.

\nop{To ensure the final causal graph $\mat{G}$ is acyclic, instead of directly modifying the edge set $\mathcal{E}_{0}$ with the (non-)existence constraints from \moduleb, the SCD algorithm used in the Causal Graph Estimator ($\S$\ref{sec.cge}) will be rerun with the constraints imposed to produce a new causal graph $\mat{G}$. Specifically, upon receiving the (non-)existence constraints, a {\em constraint matrix} $\mat{C}\in \mathbb{R}^{n\times n}$ will be constructed with $\mat{C}_{ij}=1$ if \moduleb\ suggests a causal effect from $v_{i}$ to $v_{j}$; $\mat{C}_{ij}=0$ otherwise. The most representative SCD algorithms, {\em i.e.}, PC, ES, and DirectLiNGAM, were designed to be able to incorporate such constraint matrix $\mat{C}$ in the input with observational data $\{\mat{v}_{1}$, ..., $\mat{v}_{n}\}$ for generating a causal graph that complies with the constraints in $\mat{C}$ (but to different extents), thus facilitating achieving a refined causal graph $\mat{G}$ that is also a DAG.

\vspace{0.2cm}

\noindent{\bf Remark.} Compared to the existing approaches using LLMs for causal discovery \cite{takayama2024integrating,khatibi2024alcm}, the key novelty of the proposed \method\ lies in the exploration of the \modulea\ for multi-modal enhancement of causal discovery. An algorithmic summary of the proposed workflow is in Appendix \ref{app:alg}.}

\begin{table*}[t]
\centering
\small
\setlength{\tabcolsep}{2pt}{
\begin{tabular}{l|ccccc|ccccc|ccccc}
\toprule[1.5pt]
\multirow{2}{*}{Method}       & \multicolumn{5}{c|}{\texttt{AutoMPG}}              & \multicolumn{5}{c|}{\texttt{DWDClimate}}             & \multicolumn{5}{c}{\texttt{SachsProtein}}                    \\ \cmidrule{2-16}
          & Prc $_{\uparrow}$           & F1 $_{\uparrow}$           & FPR $_{\downarrow}$           & SHD $_{\downarrow}$      & NHD $_{\downarrow}$          & Prc $_{\uparrow}$          & F1 $_{\uparrow}$             & FPR $_{\downarrow}$           &SHD $_{\downarrow}$         & NHD $_{\downarrow}$           & Prc $_{\uparrow}$          & F1 $_{\uparrow}$            & FPR $_{\downarrow}$           & SHD $_{\downarrow}$         & NHD $_{\downarrow}$ \\ \midrule
PC              & 0.11           & 0.14           & 0.40           & 8           & 0.32           & 0.14           & 0.15            & 0.20           & 9           & 0.25           &\underline{0.38}   &\textbf{0.44}   & 0.15           & 24           & 0.19 \\
Exact Search    & 0.25           & 0.30           & 0.30           & 6           & 0.24           & 0.45           & 0.58            & 0.20           &\underline{6}&\underline{0.16}& 0.18           & 0.23           & 0.26           & 31           & 0.25\\
DirectLiNGAM    & 0.11           & 0.14           & 0.40           & 8           & 0.32           & 0.16           & 0.22            & 0.33           & 10          & 0.27           & 0.27           &0.36            & 0.25           & 29           & 0.23 \\\midrule
MAC*             & -              & -              & -              & 4           & 0.16           & -              & -               & -              &\underline{6}& 0.19           & -              & -              & -              & 21           & 0.19 \\
Efficient-CDLMs &\underline{0.66}& 0.50           &\underline{0.05}& 4           & 0.16           & 0.33           & 0.33            &  0.13          & 8           & 0.22           &0.33& 0.09           &\textbf{0.02}   &\underline{20}&\underline{0.16}   \\ 
SCD-LLM  & 0.57           &\underline{0.66}& 0.15           &\underline{3}&\underline{0.12}& 0.33           & 0.22            &\underline{0.06}& 7           & 0.19           & 0.04           & 0.05           & 0.19           & 29           & 0.23 \\
ReAct           & 0.50           & 0.54           & 0.15           & 4           & 0.16           &\underline{0.60}& 0.40            &\underline{0.06}&\underline{6}&\underline{0.16}& 0.04           & 0.04           & 0.20           & 29           & 0.23 \\
LLM-KBCI        & 0.57           &\underline{0.66}& 0.15           &\underline{3}&\underline{0.12}& 0.50           & 0.40            &\underline{0.06}&\underline{6}&\underline{0.16}& 0.13           & 0.14           & 0.19           & 27           & 0.22 \\
LLM-KBCI-RA     & 0.57           & 0.55           & 0.15           &\underline{3}&\underline{0.12}&\textbf{0.75}   & \underline{0.60}&\textbf{0.03}   &\textbf{4}   &\textbf{0.11}   & 0.08           & 0.09           & 0.20           & 30           & 0.24 \\
LLM-KBCI-RE     & 0.50           & 0.61           & 0.20           & 4           & 0.16           & 0.50           & 0.40            &\underline{0.06}&\underline{6}&\underline{0.16}& 0.09           & 0.10           & 0.18           & 28           & 0.23 \\\midrule
\method         &\textbf{1.00}   &\textbf{0.88}   &\textbf{0.00}   &\textbf{1}   &\textbf{0.04}   &\textbf{0.75}   &\textbf{0.88}    &\textbf{0.03}   &\textbf{4}   &\textbf{0.11}   &\textbf{0.50}   & \underline{0.42} &\underline{0.06}&\textbf{17}   &\textbf{0.14}\\
\method-RE      & 0.57           &\underline{0.66}& 0.15           &\underline{3}&\underline{0.12}& 0.50           & 0.40            &\underline{0.06}&\underline{6}&\underline{0.16}& 0.31           & 0.31           & 0.12           & 21           & 0.17 \\ 
\bottomrule[1.5pt]
\end{tabular}

}
\caption{\label{tab:GTdatasets}
Comparison of different causal discovery methods on datasets with continuous variables. ${\uparrow}$ indicates larger score is better. ${\downarrow}$ indicates smaller score is better. * indicates numbers are adopted from the papers of the methods.
}
\end{table*}

\begin{table*}[t]
\centering
\small
\setlength{\tabcolsep}{3pt}{
\begin{tabular}{l|ccccc|ccccc}
\toprule[1.5pt]
\multirow{2}{*}{Method}               & \multicolumn{5}{c|}{\texttt{Asia}}                     & \multicolumn{5}{c}{\texttt{Child}}                   \\ \cmidrule{2-11}
          & Prc $_{\uparrow}$          & F1 $_{\uparrow}$            & FPR $_{\downarrow}$           & SHD $_{\downarrow}$        & NHD $_{\downarrow}$           & Prc $_{\uparrow}$          & F1 $_{\uparrow}$            & FPR $_{\downarrow}$           & SHD $_{\downarrow}$          & NHD $_{\downarrow}$ \\ \midrule
PC              & 0.50           & 0.50           & 0.07           & 6           & 0.09           & 0.30           & 0.34           & 0.06           & 24            & 0.06 \\
Exact Search    & 0.50           & 0.42           &\underline{0.05}& 6           & 0.09           & 0.35           & 0.28           &\textbf{0.02}   &\textbf{19}    &\textbf{0.04}    \\
DirectLiNGAM    & 0.28           & 0.36           & 0.17           & 11          & 0.17           & 0.29           & 0.36           & 0.07           & 33            & 0.08  \\\midrule
Efficient-CDLMs & 0.57           &\underline{0.53} &\underline{0.05}& 7           & 0.10           & 0.21           & 0.20           & 0.048           & 37            & 0.09    \\ 
SCD-LLM  &\underline{0.60}& 0.40           &\textbf{0.03}   &\underline{5}&\underline{0.07}&\textbf{0.56}   &\textbf{0.54}   &\textbf{0.02}   &\textbf{19}    &\textbf{0.04} \\
ReAct           & 0.40           & 0.30           &\underline{0.05}& 6           & 0.09           &\textbf{0.56}   &\textbf{0.54}   &\textbf{0.02}   &\textbf{19}    &\textbf{0.04} \\
LLM-KBCI        & 0.42           & 0.40           & 0.07           & 6           & 0.09           &\underline{0.48}&\underline{0.50}&\underline{0.03}&\underline{20} &\underline{0.05} \\
LLM-KBCI-RA     & 0.33           & 0.28           & 0.07           & 7           & 0.10           & 0.44           & 0.46           & 0.04           & 21            & 0.05 \\
LLM-KBCI-RE     & 0.28           & 0.26           & 0.08           & 7           & 0.10           & 0.40           & 0.42           & 0.04           & 22            & 0.05 \\\midrule
\method         & 0.50           & 0.42           & 0.05           & 6           & 0.09           &\underline{0.48}&\underline{0.50}&\underline{0.03}&\underline{20} & \underline{0.05} \\
\method-RE      &\textbf{0.66}   &\textbf{0.57}   &\textbf{0.03}   &\textbf{4}   &\textbf{0.06}   &\textbf{0.56}   &\textbf{0.54}   &\textbf{0.02}   &\textbf{19}    &\textbf{0.04} \\
\bottomrule[1.5pt]
\end{tabular}

}
\caption{\label{tab:bayesianDatasets}
Comparison of different causal discovery methods on datasets with discrete variables.
}
\vspace{-0.2cm}
\end{table*}

\vspace{-0.1cm}

\section{Experiments}

In this section, we first compare \method\ with SOTA methods on benchmark datasets. Then we evaluate \method\ for a root cause analysis task on real-life enterprise microservice system datasets.


\subsection{Experimental Setup}

\noindent{\bf Benchmark Datasets.} To be comprehensive, we use 5 benchmark datasets covering both continuous variables and discrete variables. For the former, following \cite{takayama2024integrating,le2024multi}, we adopt (1) \texttt{AutoMPG} \cite{quinlan1993auto}, which has five variables concerning city-cycle fuel consumption in miles per gallon, each variable has a length-392 time series; (2) \texttt{DWDClimate} \cite{mooij2016distinguishing}, which has six variables pertinent to observations from weather stations in Deutscher Wetterdienst, each variable has a length-350 time series; and (3) \texttt{SachsProtein} \cite{sachs2005causal}, which has eleven variables measuring the expression level of different proteins and phospholipids in human cells, each variable has a length-7,466 time series. For the latter, following \cite{long2023causal,jiralerspong2024efficient}, we adopt (4) \texttt{Asia} \cite{lauritzen1988local}, which has eight variables relevant to lung disease diagnosis, each variable has 1,000 discrete samples; and (5) \texttt{Child} \cite{spiegelhalter1992learning}, which has twenty variables regarding congenital heart disease in newborn babies, each variable has 1,000 discrete samples.

In \texttt{AutoMPG}, \texttt{DWDClimate}, and \texttt{SachsProtein}, ground truth causal graphs constructed by experts are available for evaluation purpose. For \texttt{Asia} and \texttt{Child}, since the observational data of the variables are sampled from a Bayesian network, the prior conditional probabilities among the variables establish the ground truth causal graphs.

\begin{table*}[t]
\centering
\small
\setlength{\tabcolsep}{1pt}{
\begin{tabular}{l|ccccc|ccccc|ccccc}
\toprule[1.5pt]
\multirow{2}{*}{Method} & \multicolumn{5}{c|}{\texttt{AutoMPG}}              & \multicolumn{5}{c|}{\texttt{DWDClimate}}             & \multicolumn{5}{c}{\texttt{SachsProtein}}                    \\ \cmidrule{2-16}
                  & Prc $_{\uparrow}$ & F1 $_{\uparrow}$  & FPR $_{\downarrow}$ & SHD $_{\downarrow}$ & NHD $_{\downarrow}$  & Prc $_{\uparrow}$ & F1 $_{\uparrow}$  & FPR $_{\downarrow}$ & SHD $_{\downarrow}$ & NHD $_{\downarrow}$  &  Prc $_{\uparrow}$ & F1 $_{\uparrow}$  & FPR $_{\downarrow}$ & SHD $_{\downarrow}$ & NHD $_{\downarrow}$ \\ \midrule
\method      & \textbf{1.00} & \textbf{0.88} & \textbf{0.00} & \textbf{1}   & \textbf{0.04}  & \textbf{0.75} & \textbf{0.88} & \textbf{0.03} & \textbf{4}   & \textbf{0.11}  & 0.50 & 0.42 & 0.06 & 17 & 0.14 \\ \midrule
(a) Iter.$\rightarrow$Single search          & 0.66 & 0.72 & 0.10 & 2   & 0.08  & 0.66 & 0.44 & \textbf{0.03} & 5   & 0.13  &  0.38 & 0.22 & 0.04 & 17 & 0.14 \\
(b) w/o Knowledge LLM          & 0.50 & 0.61 & 0.20 & 4   & 0.16  & 0.66 & 0.66 & 0.06 & \textbf{4}   & \textbf{0.11}  &  0.19 & 0.20 & 0.16 & 26 & 0.21 \\
(c) PC$\rightarrow$ES    & 0.33 & 0.42 & 0.30 & 6   & 0.24  & 0.50 & 0.50 & 0.10 & 5   & 0.13  &  0.26 & 0.28 & 0.16 & 25 & 0.20 \\
(c) PC$\rightarrow$DirectLiNGAM    & 0.16 & 0.18 & 0.25 & 6   & 0.24  & 0.16 & 0.22 & 0.23 & 9   & 0.25  &  0.19 & 0.22 & 0.20 & 27 & 0.22  \\
(d) LLM$\rightarrow$GPT4            & 0.57 & 0.66 & 0.15 & 3   & 0.12  & 0.60 & 0.54 & 0.06 & 5   & 0.13  &  \textbf{0.58} & \textbf{0.45} & \textbf{0.04} & \textbf{15} & \textbf{0.12} \\
(d) LLM$\rightarrow$Llama3.1-8B     & 0.37 & 0.46 & 0.25 & 5   & 0.20  & 0.33 & 0.22 & 0.06 & 7   & 0.19  &  0.16 & 0.18 & 0.20 & 29 & 0.23 \\
(d) LLM$\rightarrow$Llama3.1-70B    & 0.50 & 0.54 & 0.15 & 4   & 0.16  & 0.40 & 0.36 & 0.10 & 7   & 0.19  &  0.26 & 0.26 & 0.13 & 24 & 0.19 \\
(d) LLM$\rightarrow$Gemma2-9B       & 0.37 & 0.46 & 0.25 & 5   & 0.20  & 0.16 & 0.16 & 0.16 & 8   & 0.22  &  0.16 & 0.18 & 0.20 & 29 & 0.23 \\
(d) LLM$\rightarrow$Ministral-7B    & 0.60 & 0.60 & 0.10 & 4   & 0.16  & 0.33 & 0.33 & 0.13 & 7   & 0.19  &  0.21 & 0.18 & 0.10 & 25 & 0.20 \\
\bottomrule[1.5pt]
\end{tabular}

}
\caption{\label{tab: Ablation experiment}
Ablation analysis of the proposed \method\ method on benchmark datasets.
}
\vspace{-0.2cm}
\end{table*}

\vspace{0.1cm}

\noindent{\bf Baselines.} We compare \method\ with the most relevant SOTA methods on causal discovery, including (1) statistical causal discovery: Peter-Clark (PC) algorithm \cite{spirtes1991algorithm}, Exact Search (ES) algorithm \cite{yuan2013learning}, and DirectLiNGAM \cite{shimizu2011directlingam}; (2) LLM-based causal discovery that only uses LLMs to infer causal relationships: Efficient-CDLMs \cite{jiralerspong2024efficient}, which employs a BFS-based LLM prompting for efficient causal graph construction, and MAC \cite{le2024multi}, which uses Debating LLMs for building a multi-agent system; and (3) Hybrid approaches that refine an SCD causal graph by LLMs: SCD-LLM, which uses a single LLM upon the SCD output, ReAct \cite{yao2022react}, which interleaves reasoning and search tool usage when refining SCD graphs, LLM-KBCI \cite{takayama2024integrating}, which uses ZSCOT two-stage prompting for refining causal graphs.

Moreover, we apply ReAct framework for LLM-KBCI to enable alternate reasoning and tool usage and name this baseline as LLM-KBCI-RA. We also introduce Top-K Guess reasoning \cite{tian2023just} (with K=2) for verbal calibration on LLM-KBCI and name this variant as LLM-KBCI-RE. 

For our method, we consider two major variants. The first asks for a single answer from the \moduleb\ ($\S$\ref{sec.cc_agent}), {\em i.e.}, K=1 in the Top-K Guess reasoning, named as \method. The second uses K=2 in the Top-K Guess reasoning, which we name as \method-RE. Additionally, we perform extensive ablation analysis on other variants of \method\ in Table \ref{tab: Ablation experiment} to evaluate its design choices.

\vspace{0.1cm}

\noindent{\bf Implementation.} By default, we use GPT-4o mini with temperature 0.5 as the base LLM for all LLM-based and Hybrid methods, as it was found very performant by the existing works \cite{le2024multi}. For all Hybrid approaches, PC is used as the base SCD algorithm. Also, we evaluate our \method\ by switching the LLM with GPT-4, Llama-3.1-8B, Llama-3.1-70B, Mistral-7B and Gemma2-9B, and switching the SCD algorithm with ES and DirectLiNGAM in our ablation analysis. All SCD algorithms were implemented with \texttt{causal-learn} \cite{zheng2024causal}. ReAct and RAG frameworks were implemented with \texttt{LlamaIndex} \cite{LiuLlamaIndex2022}. For all baselines, we used their official code when available. Since MAC's code is unavailable, we report its results from the original paper, which are only available on \texttt{AutoMPG}, \texttt{DWDClimate}, and \texttt{SachsProtein} datasets.

\vspace{0.1cm}

\noindent{\bf Evaluation Metrics.} Following \cite{kiciman2023causal,takayama2024integrating,khatibi2024alcm,le2024multi}, we employ the widely used metrics including precision (Prc), F1-score (F1), FPR, structural Hamming distance (SHD), and normalized Hamming distance (NHD) for gauging the difference between the predicted causal graphs and the ground truth graphs. Prc and F1 measure the accuracy, thus a larger value is better. FPR, SHD, and NHD measure the errors/differences, hence a smaller value is better.

\subsection{Experimental Results}\label{sec.exp_results}




\noindent{\bf Causal Discovery.} Table \ref{tab:GTdatasets} and \ref{tab:bayesianDatasets} summarize the results on the benchmark datasets. From the tables, we have several observations: (1) Methods involving LLMs generally outperform SCD algorithms in most cases except for \texttt{SachsProtein} and \texttt{Asia} datasets which pertain to biomedicine. It demonstrates the great potential of LLMs' commonsense knowledge acquired by pre-training for the task of causal discovery, but meanwhile draws to our attention on their short of domain-specific knowledge. (2) Hybrid approaches outperform LLM-based baselines in most cases, indicating using SCD outputs as a prior for LLMs to reference has the benefits of complementing their causal effects related knowledge. (3) Baseline methods that can leverage tools for retrieving external data, {\em i.e.}, ReAct and LLM-KBCI-RA, sometimes outperform their counterparts, suggesting the potential of the 
augmented knowledge, but calling for a better way of retrieving and using 
external data. (4) The proposed \method(-RE) achieved the best overall performance considering no baseline methods consistently performs well across all datasets. In particular, \method(-RE) helps alleviate the hallucination problem of other LLM baselines on the biomedical \texttt{SachsProtein} and \texttt{Asia} datasets via data augmentation. The results highlight the challenge of the existing LLM baselines in inferring causal effects solely by meta-data ({\em i.e.}, node names, data titles) and validate the effectiveness of the proposed design of a multi-agent system for exploring multi-modality enhanced causal discovery. Finally, (5) \method-RE outperform \method\ on the Bayesian datasets \texttt{Asia} and \texttt{Child}, which could be a result of the better calibrated likelihoods by Top-K Guess reasoning for the probabilistic datasets.

\begin{figure*}[!t]
\centering
\includegraphics[width=0.99\linewidth]{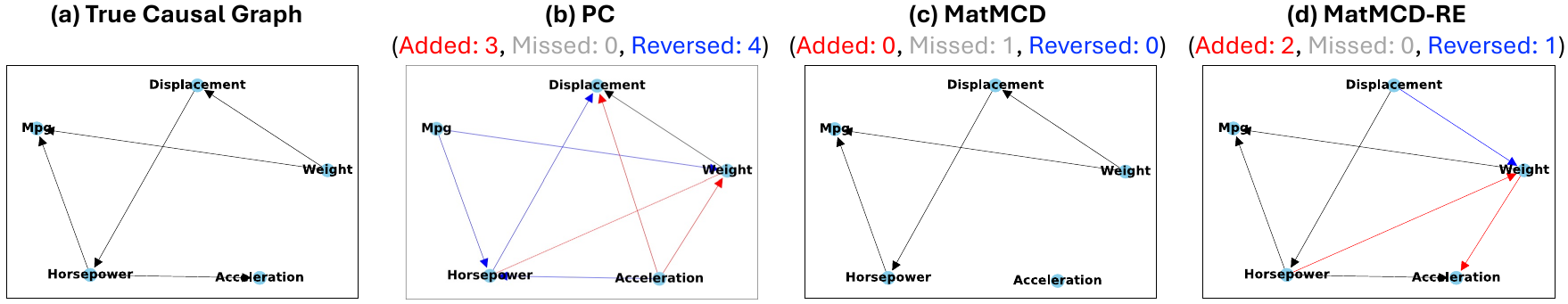}
\caption{Causal graph visualization for \texttt{AutoMPG} dataset. \textcolor{red}{Red} (\textcolor{blue}{Blue}) arrows indicate edges that are wrongly added (reversed). Each sub-caption includes the number of wrongly \textcolor{red}{added}, \textcolor{gray}{missed}, and \textcolor{blue}{reversed} edges for comparison.}\label{fig.vis1}
\end{figure*}

\begin{figure*}[!t]
\begin{minipage}[c]{0.4\textwidth}
\centering
\includegraphics[width=1.0\columnwidth]{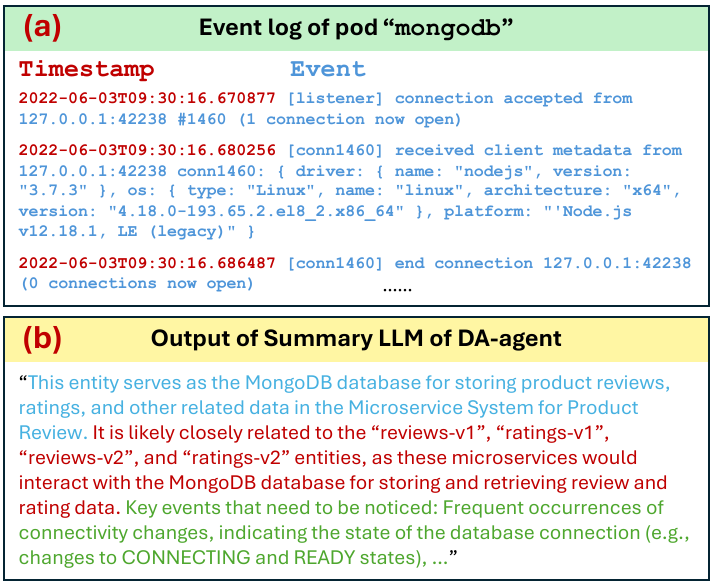}
\caption{An illustration of (a) log of events, and (b) its summary provided by \modulea.}\label{fig.log}
\end{minipage}
\hfill
\begin{minipage}[c]{0.57\textwidth}
\centering
\small
\setlength{\tabcolsep}{1.5pt}{
\begin{tabular}{l|ccccc}
\toprule[2pt]
Method                & MAP@5 $_{\uparrow}$ & MAP@10 $_{\uparrow}$  & MRR $_{\uparrow}$  & RK (P) $_{\downarrow}$ & RK (C) $_{\downarrow}$ \\\midrule
PC              & 0.0\%            & 25.0\%             & 0.14             &5                      &13 \\
Efficient-CDLMs & 0.0\%            & 0.0\%               & 0.10             &10                     &10\\
SCD-LLM  & 0.0\%            & 25.0\%              & 0.14             &5                      &13\\
ReAct           & 0.0\%            & 25.0\%              & 0.14             &5                      &12\\
LLM-KBCI        & 10.0\%           &\underline{30.0\%}   & 0.16             &4                      &13\\
LLM-KBCI-RA & 0.0\%            & 25.0\%              & 0.14             &5                      &12\\
LLM-KBCI-RE      & 10.0 \%          &\underline{30.0\%}   & 0.16             &4                      &13\\\midrule
\method         &\textbf{30.0\%}   &\textbf{55.0\%}      &\textbf{0.32}     &\textbf{2}             &\underline{7} \\
\method-RE       &\underline{20.0\%}&\textbf{55.0\%}      &\underline{0.25}  &\underline{3}          &\textbf{6}\\
\bottomrule[1.5pt]
\end{tabular}
}
\captionof{table}{Comparison of different causal discovery methods for the RCA task. RK (P) and RK (C) are the ranks of the root causes in the Product Review dataset and Cloud Computing dataset respectively, predicted by different methods.}\label{tab:LEMMA-RCA}
\end{minipage}
\end{figure*}

\vspace{0.1cm}

\noindent{\bf Ablation Study.} Table \ref{tab: Ablation experiment} presents our ablation analysis using three datasets. In the table, \method\ is our original model. In (a), we aim to study the usefulness of the iterative search in \modulea. To this end, we replaced the iterative search with a single-round search template (Appendix \ref{app:single round}). In (b), we assess the effectiveness of the Knowledge LLM in \moduleb. This is achieved by removing it from \moduleb, which is equivalent to merge Knowledge LLM and Constraint LLM into a single LLM (the template is in Appendix \ref{app.ccagent}). In (c), we evaluate how \method\ could generalize to different SCD algorithms by switching PC with ES and DirectLiNGAM. In (d), we assess the impact of different LLMs on causal discovery.

From Table \ref{tab: Ablation experiment}(a), we observe iterative search is better than single-round search, as the latter may use biased query and have difficulty in producing comprehensive augmented data. In Appendix \ref{app:DA-Agent example}, we summarized some queries iteratively generated by \modulea. From Table \ref{tab: Ablation experiment}(b), we can see including a separate Knowledge LLM for explaining causal relationships is better than merging it with Constraint LLM, suggesting their distinct roles in \moduleb. In Table \ref{tab: Ablation experiment}(c), using other SCD algorithms than PC generally degrades the performance, possibly because the constraint-based design of PC leads to better use of the constraints produced by LLMs. We also observe that, with \method, the performance of ES and DirectLiNGAM were slightly enhanced compared to the counterparts without \method\ in Table \ref{tab:GTdatasets}. Intriguingly, Table \ref{tab: Ablation experiment}(d) shows the default GPT-4o mini is the most robust across different datasets, and GPT4 is better than the open source LLMs. This is similar to the findings in \cite{le2024multi}, where MAC with GPT-4o mini performed the best. Compared to PC in Table \ref{tab:GTdatasets}, all LLM variants in Table \ref{tab: Ablation experiment}(d) improve performance, suggesting the effectiveness in leveraging the LLMs. Moreover, on the biological dataset \texttt{SachsProtein}, a larger model GPT4 further boost the causal discovery performance, likely due to its better alignment with the specific domain.

\vspace{0.1cm}

\noindent{\bf Graph Visualization.} In Fig. \ref{fig.vis1}, we investigate how \method(-RE) refines the causal graphs generated by SCD algorithms by visualizing the true causal graph of the \texttt{AutoMPG} dataset, along with the causal graphs produced by PC, \method\ and \method-RE. The visualization results on other datasets are deferred to Appendix $\S$\ref{app:Graph Visualization}. Compared to PC, \method(-RE) yields graphs that resemble the true causality better, with less wrongly added, missed, and reversed edges, suggesting the advantage of leveraging LLM agents' ability to impose multi-modal data empowered knowledge-driven constraints for screening erroneous edges. In contrast, PC alone relies solely on observational data thus is more error-prone. Moreover, LLM-provided knowledge offers precise guidance in establishing causal directions, resulting in fewer reversed edges on most of the datasets.


\subsection{Case Study: Root Cause Analysis (RCA)}
Next, we evaluate \method\ on AIOps datasets collected from Product Review (PR) and Cloud Computing (CC) microservice systems \cite{zheng2024lemma}. The PR dataset has 216 variables ({\em i.e.}, system pods), each associated with a multivariate time series containing 6 metrics (\textit{e.g.}, CPU, memory usage) of length 131,329. The CC dataset has 168 variables, each with a multivariate time series of 7 metrics and a length of 109,351. In both datasets, each variable also has a log recording its historical events. Fig.~\ref{fig.log}(a) illustrates a log snippet. Through the log lookup tool ($\S$\ref{sec_tool_preparation}), these logs can be leveraged as an additional data modality by \method\ for enhanced causal discovery. Fig.~\ref{fig.log}(b) demonstrates that the Summary LLM of \modulea\ can effectively interpret and summarize the log data.

\nop{Next, we evaluate \method\ on two AIOps datasets collected from Product Review (PR) and Cloud Computing (CC) microservice systems \cite{zheng2024lemma}. The PR dataset has 216 variables ({\em i.e.}, system pods), each variable has a multivariate time series (with 6 metrics such as CPU, memory usage, {\em etc.}) of length 131,329. The CC dataset has 168 variables, each has a multivariate time series (with 7 metrics) of length 109,351. In both datasets, each variable also has a log recording its historical events. Fig. \ref{fig.log}(a) illustrates a log snippet. Through the log lookup tool in Fig. \ref{fig.tool}, these logs can be leveraged as an additional data modality by \method\ for enhanced causal discovery. Fig. \ref{fig.log}(b) shows the Summary LLM of \modulea\ can effectively understand the log data.}

Since these datasets provide root causes of system failures identified by domain experts, but lack ground truth causal graphs, we use them to evaluate the root cause analysis (RCA) performance based on the causal graphs produced by different methods. Following \cite{wang2023interdependent}, if running a random walk with restart (RWR) on a causal graph can top-rank the root cause among all variables, it reflects the quality of the causal graph. Therefore, we use widely adopted metrics, including Mean Average Precision@K (MAP@K), with K set to 5 and 10, and Mean Reciprocal Rank (MRR), to assess the overall ranking performance on both datasets \cite{zheng2024lemma}. More details on MAP@K and MRR can be found in Appendix \ref{app:Metrics}.

\nop{Since these datasets provide root cause variables of system failure identified by domain experts, but not ground truth causal graphs, we use them for evaluating root cause analysis (RCA) performance with the causal graphs produced by different methods. Following \cite{wang2023interdependent}, if running a random walk with restart (RWR) on a causal graph can top-rank the root cause among all variables, it also reflects the good quality of the causal graph. As such, we use the widely used metrics, including Mean Average Precision@K (MAP@K), where we set K=5, 10, and Mean Reciprocal Rank (MRR) for evaluating the overall ranking performance on both datasets \cite{zheng2024lemma}. More details about MAP and MRR can be found in Appendix \ref{app:Metrics}.}

Following \cite{wang2023interdependent}, we preprocessed the datasets by filtering out irrelevant pods using an Extreme Value Theory-based approach~\cite{siffer2017anomaly} before applying causal discovery methods. Table~\ref{tab:LEMMA-RCA} summarizes the results, where we also report the specific ranking (``RK'') of the ground truth root causes for each dataset. In Table~\ref{tab:LEMMA-RCA}, ES and DirectLiNGAM are excluded as the focus is how different LLM-based methods can improve their base SCD algorithm, {\em i.e.}, PC, for the downstream task. From Table \ref{tab:LEMMA-RCA}, by leveraging the log modality, \method(-RE) significantly improves the accuracy of root cause locating on both datasets, with an 83.3\% relative improvement over the best baseline on MAP@10. This highlights the potential of integrating prevalent log data in microservice systems for RCA and AIOps.

\section{Conclusion}
In this paper, we explored multi-modal data in causal discovery via devising \method\ with tool-augmented LLMs, which has a well-crafted mechanism for integrating textual and observational data. The experiments not only validate the effectiveness of our innovative method but also set a groundwork for delving into multi-modal causal discovery.



\section{Limitations}

The research of LLM based causal discovery is an emergent area undergoing active explorations. This work pioneers the use of tool-augmented LLM agents, but the proposed method shares similar limitations of most of the existing approaches.

First, the examination of the causal relationships in a causal graph is pair-based, {\em i.e.}, each causal relationship $(v_{i}, v_{j})$ for $1 \le i < j \le n$ needs to be prompted to an LLM (in our case, \moduleb) for assessing its correctness. This may lead to scalability issues on large graphs. This is also reflected in the sizes of the benchmark datasets widely used in this area. A promising way to improve the scalability of such edge-based QA paradigm is to leverage the DAG structure of causal graphs and use breadth first search (BFS) style prompts as introduced in \cite{jiralerspong2024efficient}. The proposed \method\ is completely compatible with this prompting technique, thus has the potential to achieve an $O(n)$ complexity. Second, similar to other hybrid approaches that combine SCD algorithm and LLM agents, the proposed \method\ necessitates a base SCD algorithm that can effectively incorporates the constraints produced by LLMs. We observed this from the better results of using PC compared to use either ES or DirectLiNGAM in Table \ref{tab: Ablation experiment}. This is worth further study with a comprehensive comparison of using more constraint-based SCD algorithms and non-constraint-based variants. Third, as a knowledge-driven method, meta-data with semantically meaningful variables is needed for retrieving useful knowledge about causal relationships. For domains where variables are semantically meaningless or meta-data are private, such methods may have limited impact.

In addition to the aforementioned common limitations, this work has several specific limitations. First, this work uses web data and logs as the augmented data modality of focus, more modalities such as codes, images and audio signals in certain domains remain further study. Second, 
tools beyond the web search APIs and log lookup APIs in the proposed \modulea, such as Wikipedia APIs and code lookup APIs, are worth exploring. The toolkit in the \modulea\ has the flexibility and extensibility to support further research.


\bibliography{ref}

\begin{thebibliography}{57}
\providecommand{\natexlab}[1]{#1}

\bibitem[{Achiam et~al.(2023)Achiam, Adler, Agarwal, Ahmad, Akkaya, Aleman, Almeida, Altenschmidt, Altman, Anadkat et~al.}]{achiam2023gpt}
Josh Achiam, Steven Adler, Sandhini Agarwal, Lama Ahmad, Ilge Akkaya, Florencia~Leoni Aleman, Diogo Almeida, Janko Altenschmidt, Sam Altman, Shyamal Anadkat, et~al. 2023.
\newblock Gpt-4 technical report.
\newblock \emph{arXiv preprint arXiv:2303.08774}.

\bibitem[{Ban et~al.(2023)Ban, Chen, Lyu, Wang, and Chen}]{ban2023causal}
Taiyu Ban, Lyuzhou Chen, Derui Lyu, Xiangyu Wang, and Huanhuan Chen. 2023.
\newblock Causal structure learning supervised by large language model.
\newblock \emph{arXiv preprint arXiv:2311.11689}.

\bibitem[{Brown et~al.(2020)Brown, Mann, Ryder, Subbiah, Kaplan, Dhariwal, Neelakantan, Shyam, Sastry, Askell et~al.}]{brown2020languagemodelsfewshotlearners}
Tom Brown, Benjamin Mann, Nick Ryder, Melanie Subbiah, Jared~D Kaplan, Prafulla Dhariwal, Arvind Neelakantan, Pranav Shyam, Girish Sastry, Amanda Askell, et~al. 2020.
\newblock Language models are few-shot learners.
\newblock \emph{NeurIPS}.

\bibitem[{Chen et~al.(2024{\natexlab{a}})Chen, Peng, Chen, Wang, Xu, Zeng, Zhao, Zhao, Qiao, and Lu}]{chen2024causal}
Sirui Chen, Bo~Peng, Meiqi Chen, Ruiqi Wang, Mengying Xu, Xingyu Zeng, Rui Zhao, Shengjie Zhao, Yu~Qiao, and Chaochao Lu. 2024{\natexlab{a}}.
\newblock Causal evaluation of language models.
\newblock \emph{arXiv preprint arXiv:2405.00622}.

\bibitem[{Chen et~al.(2024{\natexlab{b}})Chen, Arkin, Zhang, Roy, and Fan}]{chen2024scalable}
Yongchao Chen, Jacob Arkin, Yang Zhang, Nicholas Roy, and Chuchu Fan. 2024{\natexlab{b}}.
\newblock Scalable multi-robot collaboration with large language models: Centralized or decentralized systems?
\newblock In \emph{ICRA}.

\bibitem[{Chickering(2002)}]{chickering2002optimal}
David~Maxwell Chickering. 2002.
\newblock Optimal structure identification with greedy search.
\newblock \emph{J. Mach. Learn. Res.}, 3(Nov):507--554.

\bibitem[{Hoyer et~al.(2008)Hoyer, Janzing, Mooij, Peters, and Sch{\"o}lkopf}]{hoyer2008nonlinear}
Patrik Hoyer, Dominik Janzing, Joris~M Mooij, Jonas Peters, and Bernhard Sch{\"o}lkopf. 2008.
\newblock Nonlinear causal discovery with additive noise models.
\newblock \emph{NeurIPS}.

\bibitem[{Huang et~al.(2018)Huang, Zhang, Lin, Sch{\"o}lkopf, and Glymour}]{huang2018generalized}
Biwei Huang, Kun Zhang, Yizhu Lin, Bernhard Sch{\"o}lkopf, and Clark Glymour. 2018.
\newblock Generalized score functions for causal discovery.
\newblock In \emph{KDD}.

\bibitem[{Jiralerspong et~al.(2024)Jiralerspong, Chen, More, Shah, and Bengio}]{jiralerspong2024efficient}
Thomas Jiralerspong, Xiaoyin Chen, Yash More, Vedant Shah, and Yoshua Bengio. 2024.
\newblock Efficient causal graph discovery using large language models.
\newblock \emph{arXiv preprint arXiv:2402.01207}.

\bibitem[{Karpas et~al.(2022)Karpas, Abend, Belinkov, Lenz, Lieber, Ratner, Shoham, Bata, Levine, Leyton-Brown et~al.}]{karpas2022mrkl}
Ehud Karpas, Omri Abend, Yonatan Belinkov, Barak Lenz, Opher Lieber, Nir Ratner, Yoav Shoham, Hofit Bata, Yoav Levine, Kevin Leyton-Brown, et~al. 2022.
\newblock Mrkl systems: A modular, neuro-symbolic architecture that combines large language models, external knowledge sources and discrete reasoning.
\newblock \emph{arXiv preprint arXiv:2205.00445}.

\bibitem[{Khatibi et~al.(2024)Khatibi, Abbasian, Yang, Azimi, and Rahmani}]{khatibi2024alcm}
Elahe Khatibi, Mahyar Abbasian, Zhongqi Yang, Iman Azimi, and Amir~M Rahmani. 2024.
\newblock Alcm: Autonomous llm-augmented causal discovery framework.
\newblock \emph{arXiv preprint arXiv:2405.01744}.

\bibitem[{K{\i}c{\i}man et~al.(2023)K{\i}c{\i}man, Ness, Sharma, and Tan}]{kiciman2023causal}
Emre K{\i}c{\i}man, Robert Ness, Amit Sharma, and Chenhao Tan. 2023.
\newblock Causal reasoning and large language models: Opening a new frontier for causality.
\newblock \emph{arXiv preprint arXiv:2305.00050}.

\bibitem[{Kojima et~al.(2022)Kojima, Gu, Reid, Matsuo, and Iwasawa}]{kojima2022large}
Takeshi Kojima, Shixiang~Shane Gu, Machel Reid, Yutaka Matsuo, and Yusuke Iwasawa. 2022.
\newblock Large language models are zero-shot reasoners.
\newblock \emph{NeurIPS}.

\bibitem[{Lauritzen and Spiegelhalter(1988)}]{lauritzen1988local}
Steffen~L Lauritzen and David~J Spiegelhalter. 1988.
\newblock Local computations with probabilities on graphical structures and their application to expert systems.
\newblock \emph{J. R. Stat. Soc. Ser. B Methodol.}, 50(2):157--194.

\bibitem[{Le et~al.(2024)Le, Xia, and Chen}]{le2024multi}
Hao~Duong Le, Xin Xia, and Zhang Chen. 2024.
\newblock Multi-agent causal discovery using large language models.
\newblock \emph{arXiv preprint arXiv:2407.15073}.

\bibitem[{Lewis et~al.(2020)Lewis, Perez, Piktus, Petroni, Karpukhin, Goyal, K{\"u}ttler, Lewis, Yih, Rockt{\"a}schel et~al.}]{lewis2020retrieval}
Patrick Lewis, Ethan Perez, Aleksandra Piktus, Fabio Petroni, Vladimir Karpukhin, Naman Goyal, Heinrich K{\"u}ttler, Mike Lewis, Wen-tau Yih, Tim Rockt{\"a}schel, et~al. 2020.
\newblock Retrieval-augmented generation for knowledge-intensive nlp tasks.
\newblock \emph{NeurIPS}.

\bibitem[{Li et~al.(2023)Li, Hammoud, Itani, Khizbullin, and Ghanem}]{li2023camel}
Guohao Li, Hasan Hammoud, Hani Itani, Dmitrii Khizbullin, and Bernard Ghanem. 2023.
\newblock Camel: Communicative agents for" mind" exploration of large language model society.
\newblock \emph{NeurIPS}.

\bibitem[{Liang et~al.(2023)Liang, He, Jiao, Wang, Wang, Wang, Yang, Shi, and Tu}]{liang2023encouraging}
Tian Liang, Zhiwei He, Wenxiang Jiao, Xing Wang, Yan Wang, Rui Wang, Yujiu Yang, Shuming Shi, and Zhaopeng Tu. 2023.
\newblock Encouraging divergent thinking in large language models through multi-agent debate.
\newblock \emph{arXiv preprint arXiv:2305.19118}.

\bibitem[{Liu(2022)}]{LiuLlamaIndex2022}
Jerry Liu. 2022.
\newblock \href {https://github.com/jerryjliu/llama_index} {Llamaindex}.

\bibitem[{Long et~al.(2023)Long, Pich{\'e}, Zantedeschi, Schuster, and Drouin}]{long2023causal}
Stephanie Long, Alexandre Pich{\'e}, Valentina Zantedeschi, Tibor Schuster, and Alexandre Drouin. 2023.
\newblock Causal discovery with language models as imperfect experts.
\newblock \emph{arXiv preprint arXiv:2307.02390}.

\bibitem[{Madaan et~al.(2024)Madaan, Tandon, Gupta, Hallinan, Gao, Wiegreffe, Alon, Dziri, Prabhumoye, Yang et~al.}]{madaan2024self}
Aman Madaan, Niket Tandon, Prakhar Gupta, Skyler Hallinan, Luyu Gao, Sarah Wiegreffe, Uri Alon, Nouha Dziri, Shrimai Prabhumoye, Yiming Yang, et~al. 2024.
\newblock Self-refine: Iterative refinement with self-feedback.
\newblock \emph{NeurIPS}.

\bibitem[{Mooij et~al.(2016)Mooij, Peters, Janzing, Zscheischler, and Sch{\"o}lkopf}]{mooij2016distinguishing}
Joris~M Mooij, Jonas Peters, Dominik Janzing, Jakob Zscheischler, and Bernhard Sch{\"o}lkopf. 2016.
\newblock Distinguishing cause from effect using observational data: methods and benchmarks.
\newblock \emph{J. Mach. Learn. Res.}, 17(32):1--102.

\bibitem[{Nguyen et~al.(2023)Nguyen, Tong, Madan, Bengio, and Liu}]{nguyen2023causal}
Trang Nguyen, Alexander Tong, Kanika Madan, Yoshua Bengio, and Dianbo Liu. 2023.
\newblock Causal inference in gene regulatory networks with gflownet: Towards scalability in large systems.
\newblock \emph{arXiv preprint arXiv:2310.03579}.

\bibitem[{OpenAI(2024)}]{function_calling_2024}
OpenAI. 2024.
\newblock Function calling guide.
\newblock {URL}: https://platform.openai.com/docs/guides/function-calling.
\newblock Accessed: 2024-12-10.

\bibitem[{Qian et~al.(2023)Qian, Cong, Yang, Chen, Su, Xu, Liu, and Sun}]{qian2023communicative}
Chen Qian, Xin Cong, Cheng Yang, Weize Chen, Yusheng Su, Juyuan Xu, Zhiyuan Liu, and Maosong Sun. 2023.
\newblock Communicative agents for software development.
\newblock \emph{arXiv preprint arXiv:2307.07924}.

\bibitem[{Quinlan(1993)}]{quinlan1993auto}
Ross Quinlan. 1993.
\newblock Auto mpg.
\newblock UCI Machine Learning Repository.
\newblock {DOI}: https://doi.org/10.24432/C5859H.

\bibitem[{Rolland et~al.(2022)Rolland, Cevher, Kleindessner, Russell, Janzing, Sch{\"o}lkopf, and Locatello}]{rolland2022score}
Paul Rolland, Volkan Cevher, Matth{\"a}us Kleindessner, Chris Russell, Dominik Janzing, Bernhard Sch{\"o}lkopf, and Francesco Locatello. 2022.
\newblock Score matching enables causal discovery of nonlinear additive noise models.
\newblock In \emph{ICML}.

\bibitem[{Sachs et~al.(2005)Sachs, Perez, Pe'er, Lauffenburger, and Nolan}]{sachs2005causal}
Karen Sachs, Omar Perez, Dana Pe'er, Douglas~A Lauffenburger, and Garry~P Nolan. 2005.
\newblock Causal protein-signaling networks derived from multiparameter single-cell data.
\newblock \emph{Science}, 308(5721):523--529.

\bibitem[{Schick et~al.(2024)Schick, Dwivedi-Yu, Dess{\`\i}, Raileanu, Lomeli, Hambro, Zettlemoyer, Cancedda, and Scialom}]{schick2024toolformer}
Timo Schick, Jane Dwivedi-Yu, Roberto Dess{\`\i}, Roberta Raileanu, Maria Lomeli, Eric Hambro, Luke Zettlemoyer, Nicola Cancedda, and Thomas Scialom. 2024.
\newblock Toolformer: Language models can teach themselves to use tools.
\newblock \emph{NeurIPS}.

\bibitem[{Shen et~al.(2024)Shen, Song, Tan, Li, Lu, and Zhuang}]{shen2024hugginggpt}
Yongliang Shen, Kaitao Song, Xu~Tan, Dongsheng Li, Weiming Lu, and Yueting Zhuang. 2024.
\newblock Hugginggpt: Solving ai tasks with chatgpt and its friends in hugging face.
\newblock \emph{NeurIPS}.

\bibitem[{Shimizu et~al.(2006)Shimizu, Hoyer, Hyv{\"a}rinen, Kerminen, and Jordan}]{shimizu2006linear}
Shohei Shimizu, Patrik~O Hoyer, Aapo Hyv{\"a}rinen, Antti Kerminen, and Michael Jordan. 2006.
\newblock A linear non-gaussian acyclic model for causal discovery.
\newblock \emph{J. Mach. Learn. Res.}, 7(10).

\bibitem[{Shimizu et~al.(2011)Shimizu, Inazumi, Sogawa, Hyvarinen, Kawahara, Washio, Hoyer, Bollen, and Hoyer}]{shimizu2011directlingam}
Shohei Shimizu, Takanori Inazumi, Yasuhiro Sogawa, Aapo Hyvarinen, Yoshinobu Kawahara, Takashi Washio, Patrik~O Hoyer, Kenneth Bollen, and Patrik Hoyer. 2011.
\newblock Directlingam: A direct method for learning a linear non-gaussian structural equation model.
\newblock \emph{J. Mach. Learn. Res.}, 12(Apr):1225--1248.

\bibitem[{Shinn et~al.(2024)Shinn, Cassano, Gopinath, Narasimhan, and Yao}]{shinn2024reflexion}
Noah Shinn, Federico Cassano, Ashwin Gopinath, Karthik Narasimhan, and Shunyu Yao. 2024.
\newblock Reflexion: Language agents with verbal reinforcement learning.
\newblock \emph{NeurIPS}.

\bibitem[{Siffer et~al.(2017)Siffer, Fouque, Termier, and Largouet}]{siffer2017anomaly}
Alban Siffer, Pierre-Alain Fouque, Alexandre Termier, and Christine Largouet. 2017.
\newblock Anomaly detection in streams with extreme value theory.
\newblock In \emph{KDD}.

\bibitem[{Silander and Myllym{\"a}ki(2006)}]{silander2006simple}
Tomi Silander and Petri Myllym{\"a}ki. 2006.
\newblock A simple approach for finding the globally optimal bayesian network structure.
\newblock In \emph{UAI}.

\bibitem[{Spiegelhalter(1992)}]{spiegelhalter1992learning}
David~J Spiegelhalter. 1992.
\newblock Learning in probabilistic expert systems.
\newblock \emph{Bayesian statistics}, 4:447--465.

\bibitem[{Spirtes and Glymour(1991)}]{spirtes1991algorithm}
Peter Spirtes and Clark Glymour. 1991.
\newblock An algorithm for fast recovery of sparse causal graphs.
\newblock \emph{Soc. Sci. Comput. Rev.}, 9(1):62--72.

\bibitem[{Takayama et~al.(2024)Takayama, Okuda, Pham, Ikenoue, Fukuma, Shimizu, and Sannai}]{takayama2024integrating}
Masayuki Takayama, Tadahisa Okuda, Thong Pham, Tatsuyoshi Ikenoue, Shingo Fukuma, Shohei Shimizu, and Akiyoshi Sannai. 2024.
\newblock Integrating large language models in causal discovery: A statistical causal approach.
\newblock \emph{arXiv preprint arXiv:2402.01454}.

\bibitem[{Tian et~al.(2023)Tian, Mitchell, Zhou, Sharma, Rafailov, Yao, Finn, and Manning}]{tian2023just}
Katherine Tian, Eric Mitchell, Allan Zhou, Archit Sharma, Rafael Rafailov, Huaxiu Yao, Chelsea Finn, and Christopher~D Manning. 2023.
\newblock Just ask for calibration: Strategies for eliciting calibrated confidence scores from language models fine-tuned with human feedback.
\newblock In \emph{EMNLP}.

\bibitem[{Tu et~al.(2019)Tu, Zhang, Bertilson, Kjellstrom, and Zhang}]{tu2019neuropathic}
Ruibo Tu, Kun Zhang, Bo~Bertilson, Hedvig Kjellstrom, and Cheng Zhang. 2019.
\newblock Neuropathic pain diagnosis simulator for causal discovery algorithm evaluation.
\newblock \emph{NeurIPS}.

\bibitem[{Tu et~al.(2022)Tu, Zhang, Kjellstrom, and Zhang}]{tu2022optimal}
Ruibo Tu, Kun Zhang, Hedvig Kjellstrom, and Cheng Zhang. 2022.
\newblock Optimal transport for causal discovery.
\newblock In \emph{ICLR}.

\bibitem[{Vashishtha et~al.(2023)Vashishtha, Reddy, Kumar, Bachu, Balasubramanian, and Sharma}]{vashishtha2023causal}
Aniket Vashishtha, Abbavaram~Gowtham Reddy, Abhinav Kumar, Saketh Bachu, Vineeth~N Balasubramanian, and Amit Sharma. 2023.
\newblock Causal inference using llm-guided discovery.
\newblock \emph{arXiv preprint arXiv:2310.15117}.

\bibitem[{Wang et~al.(2023{\natexlab{a}})Wang, Chen, Fu, Liu, and Chen}]{wang2023incremental}
Dongjie Wang, Zhengzhang Chen, Yanjie Fu, Yanchi Liu, and Haifeng Chen. 2023{\natexlab{a}}.
\newblock Incremental causal graph learning for online root cause analysis.
\newblock In \emph{KDD}.

\bibitem[{Wang et~al.(2023{\natexlab{b}})Wang, Chen, Ni, Tong, Wang, Fu, and Chen}]{wang2023interdependent}
Dongjie Wang, Zhengzhang Chen, Jingchao Ni, Liang Tong, Zheng Wang, Yanjie Fu, and Haifeng Chen. 2023{\natexlab{b}}.
\newblock Interdependent causal networks for root cause localization.
\newblock In \emph{KDD}.

\bibitem[{Wang et~al.(2024)Wang, Feng, He, Tan, Han, and Tsvetkov}]{wang2024can}
Heng Wang, Shangbin Feng, Tianxing He, Zhaoxuan Tan, Xiaochuang Han, and Yulia Tsvetkov. 2024.
\newblock Can language models solve graph problems in natural language?
\newblock \emph{NeurIPS}.

\bibitem[{Wei et~al.(2022)Wei, Wang, Schuurmans, Bosma, Xia, Chi, Le, Zhou et~al.}]{wei2022chain}
Jason Wei, Xuezhi Wang, Dale Schuurmans, Maarten Bosma, Fei Xia, Ed~Chi, Quoc~V Le, Denny Zhou, et~al. 2022.
\newblock Chain-of-thought prompting elicits reasoning in large language models.
\newblock \emph{NeurIPS}.

\bibitem[{Xie et~al.(2020)Xie, Cai, Huang, Glymour, Hao, and Zhang}]{xie2020generalized}
Feng Xie, Ruichu Cai, Biwei Huang, Clark Glymour, Zhifeng Hao, and Kun Zhang. 2020.
\newblock Generalized independent noise condition for estimating latent variable causal graphs.
\newblock \emph{NeurIPS}.

\bibitem[{Xiong et~al.(2023)Xiong, Ding, Cao, Liu, and Qin}]{xiong2023examining}
Kai Xiong, Xiao Ding, Yixin Cao, Ting Liu, and Bing Qin. 2023.
\newblock Examining inter-consistency of large language models collaboration: An in-depth analysis via debate.
\newblock \emph{arXiv preprint arXiv:2305.11595}.

\bibitem[{Xiong et~al.(2024)Xiong, Hu, Lu, LI, Fu, He, and Hooi}]{xiong2024can}
Miao Xiong, Zhiyuan Hu, Xinyang Lu, YIFEI LI, Jie Fu, Junxian He, and Bryan Hooi. 2024.
\newblock Can llms express their uncertainty? an empirical evaluation of confidence elicitation in llms.
\newblock In \emph{ICLR}.

\bibitem[{Yao et~al.(2022)Yao, Zhao, Yu, Du, Shafran, Narasimhan, and Cao}]{yao2022react}
Shunyu Yao, Jeffrey Zhao, Dian Yu, Nan Du, Izhak Shafran, Karthik Narasimhan, and Yuan Cao. 2022.
\newblock React: Synergizing reasoning and acting in language models.
\newblock \emph{arXiv preprint arXiv:2210.03629}.

\bibitem[{Yu et~al.(2025)Yu, Chen, Xiong, Wu, Li, Chen, Liu, and Pan}]{yu2025causaleval}
Longxuan Yu, Delin Chen, Siheng Xiong, Qingyang Wu, Dawei Li, Zhikai Chen, Xiaoze Liu, and Liangming Pan. 2025.
\newblock Causaleval: Towards better causal reasoning in language models.
\newblock In \emph{NAACL-HLT}.

\bibitem[{Yuan and Malone(2013)}]{yuan2013learning}
Changhe Yuan and Brandon Malone. 2013.
\newblock Learning optimal bayesian networks: A shortest path perspective.
\newblock \emph{J. Artif. Intell. Res.}, 48:23--65.

\bibitem[{Ze{\v{c}}evi{\'c} et~al.(2023)Ze{\v{c}}evi{\'c}, Willig, Dhami, and Kersting}]{zevcevic2023causal}
Matej Ze{\v{c}}evi{\'c}, Moritz Willig, Devendra~Singh Dhami, and Kristian Kersting. 2023.
\newblock Causal parrots: Large language models may talk causality but are not causal.
\newblock \emph{arXiv preprint arXiv:2308.13067}.

\bibitem[{Zhao et~al.(2023)Zhao, Wang, Zhang, Jin, Zhu, Chen, and Xie}]{zhao2023competeai}
Qinlin Zhao, Jindong Wang, Yixuan Zhang, Yiqiao Jin, Kaijie Zhu, Hao Chen, and Xing Xie. 2023.
\newblock Competeai: Understanding the competition behaviors in large language model-based agents.
\newblock \emph{arXiv preprint arXiv:2310.17512}.

\bibitem[{Zheng et~al.(2024{\natexlab{a}})Zheng, Chen, Wang, Deng, Matsuoka, and Chen}]{zheng2024lemma}
Lecheng Zheng, Zhengzhang Chen, Dongjie Wang, Chengyuan Deng, Reon Matsuoka, and Haifeng Chen. 2024{\natexlab{a}}.
\newblock Lemma-rca: A large multi-modal multi-domain dataset for root cause analysis.
\newblock \emph{arXiv preprint arXiv:2406.05375}.

\bibitem[{Zheng et~al.(2018)Zheng, Aragam, Ravikumar, and Xing}]{zheng2018dags}
Xun Zheng, Bryon Aragam, Pradeep~K Ravikumar, and Eric~P Xing. 2018.
\newblock Dags with no tears: Continuous optimization for structure learning.
\newblock \emph{NeurIPS}.

\bibitem[{Zheng et~al.(2024{\natexlab{b}})Zheng, Huang, Chen, Ramsey, Gong, Cai, Shimizu, Spirtes, and Zhang}]{zheng2024causal}
Yujia Zheng, Biwei Huang, Wei Chen, Joseph Ramsey, Mingming Gong, Ruichu Cai, Shohei Shimizu, Peter Spirtes, and Kun Zhang. 2024{\natexlab{b}}.
\newblock Causal-learn: Causal discovery in python.
\newblock \emph{J. Mach. Learn. Res.}, 25(60):1--8.

\end{thebibliography}

\clearpage
\onecolumn
\appendix

\section{Algorithm}\label{app:alg}

The algorithm of \method\ is summarized in Algorithm \ref{alg.inference}. The notations are consistent with $\S$\ref{sec.method}.
\begin{algorithm}[!h]
\DontPrintSemicolon
\SetNoFillComment
\KwIn{(1) Observational Data: $\{\mat{v}_{1}, ..., \mat{v}_{n}\}$, where $\mat{v}_{i}=\{v_{i1}, ..., v_{im}\}$ includes $m$ samples for variable $v_{i}$; (2) Meta-Data $\mathcal{D}=\{\mathsf{s}, \mathcal{Z}\}$, where $\mathsf{s}$ is a descriptive title of the dataset and $\mathcal{Z}=\{\mathsf{z}_{1}, ..., \mathsf{z}_{n}\}$ includes the descriptive name $\mathsf{z}_{i}$ of each variable $v_{i}$ in $\mathcal{V}$; (3) An SCD algorithm $f_{\text{SCD}}(\cdot)$; (4) The number of guesses $K$ in Top-K Guess prompt for eliciting verbal confidence in Constraint LLM.}
\KwOut{Refined causal graph $\mat{G}$}

\BlankLine

\tcc{Causal graph estimator ($\S$\ref{sec.cge})}
Initial Causal Graph $\mat{G}_{0}\leftarrow\texttt{CausalGraphEstimator}(f_{\text{SCD}}, \{\mat{v}_{1}, ..., \mat{v}_{n}\})$\;
\tcc{\modulea\ ($\S$\ref{sec.data_agent})}
Initialize call history memory $\mathcal{C}\leftarrow\emptyset$\;
Initialize retrieved data memory $\mathcal{R}\leftarrow\emptyset$\;
Query $\leftarrow\texttt{Search-LLM}(\mathsf{s}, \mathcal{Z}, \mathcal{C})$\tcp*{Generate a search query after checking memory $\mathcal{C}$}
\While{Query $\neq$ ``No query needed''}
{
    \tcc{Iterative search}
    $\mathcal{R}\leftarrow\texttt{Toolkit}(\text{Query})$\tcp*{Add retrieved data to memory $\mathcal{R}$}
    $\mathcal{C}\leftarrow\text{Query}$\tcp*{Add search query to memory $\mathcal{C}$}
    Query $\leftarrow\texttt{Search-LLM}(\mathsf{s}, \mathcal{Z}, \mathcal{C})$\tcp*{Update search query after checking memory $\mathcal{C}$}
}
ContextualData $\leftarrow\texttt{Summary-LLM}(\mathsf{s}, \mathcal{Z}, \mathcal{R})$\;
\tcc{\moduleb\ ($\S$\ref{sec.cc_agent})}
Initialize constraint matrix $\mat{C}$\;
\For{Each pair of variables $(v_{i}, v_{j})$}{
    $\text{Explanation}_{i,j}\leftarrow\texttt{Knowledge-LLM}(\mat{G}_{0}, v_{i}, v_{j}, \text{ContextualData})$\;
    $\{(\text{Constraint}_{k}, \text{Confidence}_{k})\}_{k=1}^{K}\leftarrow\texttt{Constraint-LLM}(\text{Explanation}_{i,j})$\tcp*{Top-K guesses}
    $k_{*}\leftarrow\mathrm{argmax}_{k}\text{Confidence}_{k}$\;
    $\mat{C}_{i,j}\leftarrow\text{Constraint}_{k_{*}}$\tcp*{Update constraint matrix with $\text{Constraint}_{k_{*}}$ for $(v_{i}, v_{j})$}
}
\tcc{Causal Graph Refiner ($\S$\ref{sec.cgr})}
Refined causal graph $\mat{G}\leftarrow\texttt{CausalGraphRefiner}(f_{\text{SCD}}, \{\mat{v}_{1}, ..., \mat{v}_{n}\}, \mat{C})$

\caption{Multi-Agent with Tool-Augmented LLMs for Multi-Modality Enhanced Causal Discovery (\method)}\label{alg.inference}
\end{algorithm}

\section{Experimental Details}

\subsection{Implementation Details}

For PC algorithm \cite{spirtes1991algorithm}, Fisher’s Z test was used in its conditional independence tests. For Exact Search \cite{yuan2013learning}, its k-cycle heuristic was enabled with $k=2$ and its max parents was limited to 2. For DirectLiNGAM \cite{shimizu2011directlingam}, soft prior knowledge integration was adopted to circumvent its potential conflicts with LLM generated constraints. For LLMs, we set the temperature to 0.5 by default. However, certain LLMs ({\em e.g.}, Llama 3.1, Mistral) may produce outputs that do not adhere to the prompt requirements. In this case, we attempt tuning the temperature to 0.7 and rerunning multiple times. 
For Efficient-CDLMs \cite{jiralerspong2024efficient}, it may produce different results across multiple runs. Thus we reported its best performance of 10 runs on each dataset.

For the implementation of RAG, \texttt{LlamaIndex} was used with \texttt{text-embedding-ada-002} for document chunk indexing. For the web search tool in \modulea, Google search API via Serper\footnote{\url{https://serper.dev/}} was adopted. If the returned information is insufficient or incomplete, Tavily\footnote{\url{https://tavily.com/}} was employed as a supplement. By consolidating information from multiple sources, we ensure a comprehensive and diverse context, aiding causal discovery and LLM prompts. For the log lookup tool, template-based logs were used to systematically record events. Specifically, events were capped by 10 occurrences with random sampling.

\subsection{Evaluation Metrics in RCA}\label{app:Metrics}

To evaluate the performance of RCA, following \cite{wang2023interdependent,zheng2024lemma}, we use the following widely-used metrics.

\vspace{0.1cm}

\noindent{\textbf{Precision@K (PR@K)}} measures the probability that the Top-$K$ predicted root cause are true, which is defined as
$$
\mathrm{PR@K}=\frac{1}{|\mathbb{A}|}\sum_{a\in\mathbb{A}}\frac{\sum_{i<K}R_a(i)\in V_a}{\mathrm{min}(K,|V_a|)}
$$
where $\mathbb{A}$ is the set of system faults, $a$ is one fault in $\mathbb{A}$, $V_a$ is the set of true root causes of fault $a$, $R_a$ is the set of predicted root causes of fault $a$, and $R_{a}(i)$ is the $i$-th ranked prediction in $R_a$.

\vspace{0.1cm}

\noindent{\textbf{Mean Average Precision@K (MAP@K)}} assesses the model performance in the top-$K$ predicted root causes from the overall perspective, which is defined as following.
$$
\mathrm{MAP@K}=\frac{1}{K|\mathbb{A}|}\sum_{a\in\mathbb{A}}\sum_{1\le j\le K}\mathrm{PR@j}
$$
where a higher value indicates better performance.

\vspace{0.1cm}

\noindent{\textbf{Mean Reciprocal Rank (MRR)}} measures the ranking capability of models. A higher MRR indicates that the predicted root causes tend to appear earlier in the ranking. MRR is defined as following.
$$
\mathrm{MRR}=\frac{1}{|\mathbb{A}|}\sum_{a\in \mathbb{A}}\frac{1}{\mathrm{rank}_{R_a}}
$$
where $\mathrm{rank}_{R_a}$ is the rank of the first correctly predicted root cause of system fault $a$.

\subsection{Additional Graph Visualizations}\label{app:Graph Visualization}

\begin{figure}[!t]
\centering
\includegraphics[width=0.8\linewidth]{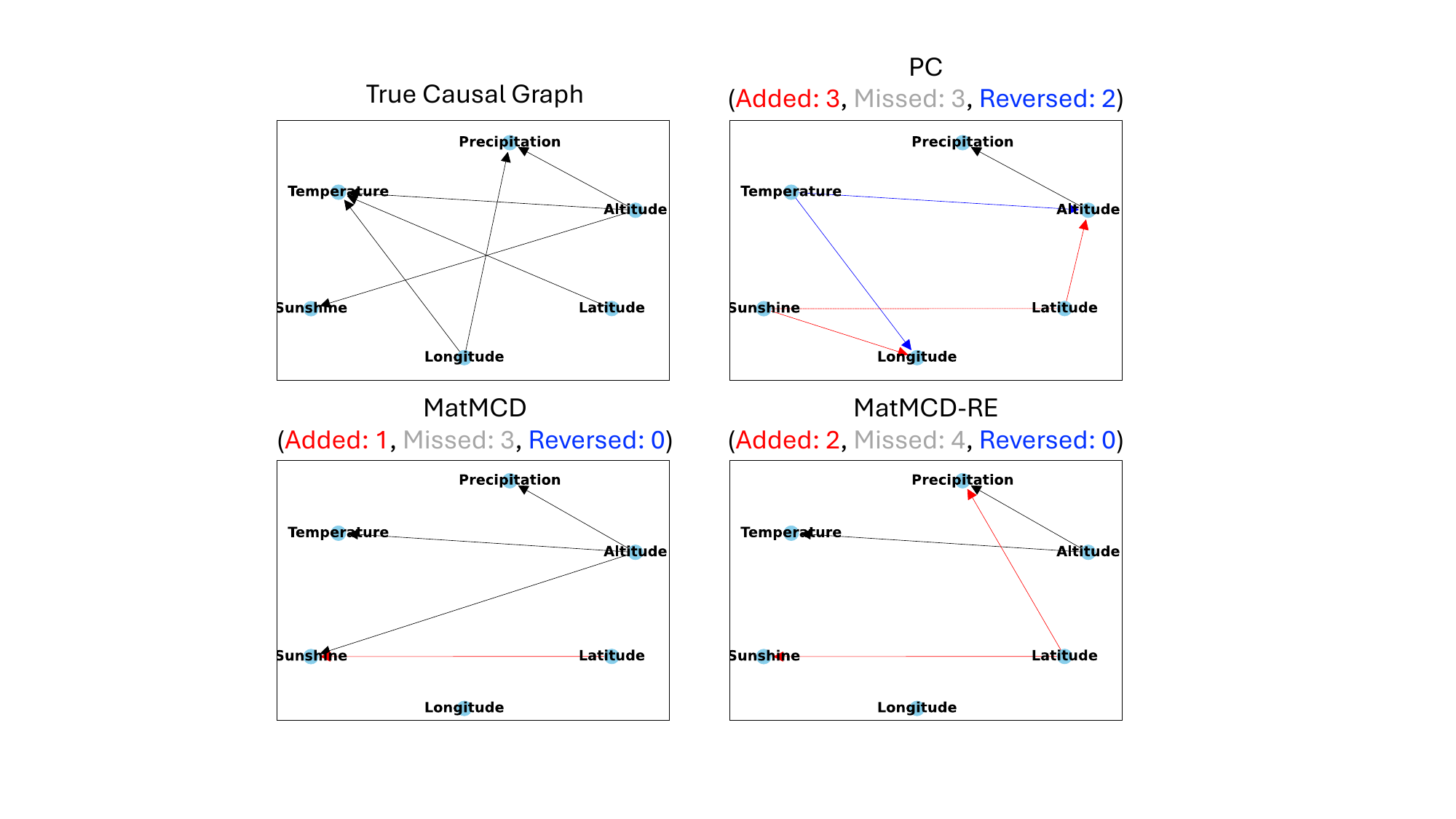}
\caption{Causal Graph visualization for \texttt{DWDClimate} dataset.}\label{fig.vis2}
\end{figure}

\begin{figure}[!t]
\centering
\includegraphics[width=0.8\linewidth]{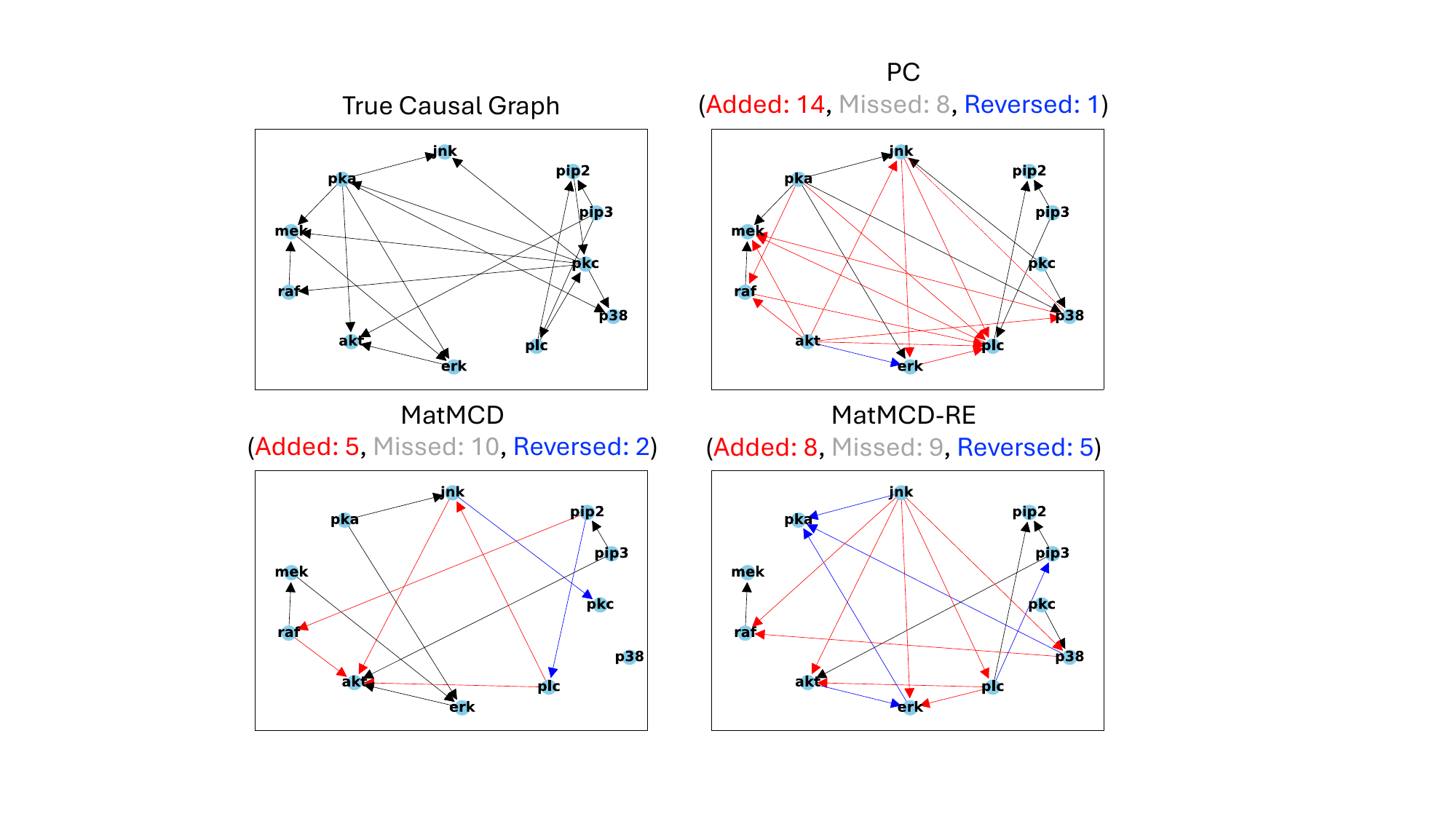}
\caption{Causal Graph visualization for \texttt{SachsProtein} dataset.}\label{fig.vis3}
\end{figure}

\begin{figure}[!t]
\centering
\includegraphics[width=0.8\linewidth]{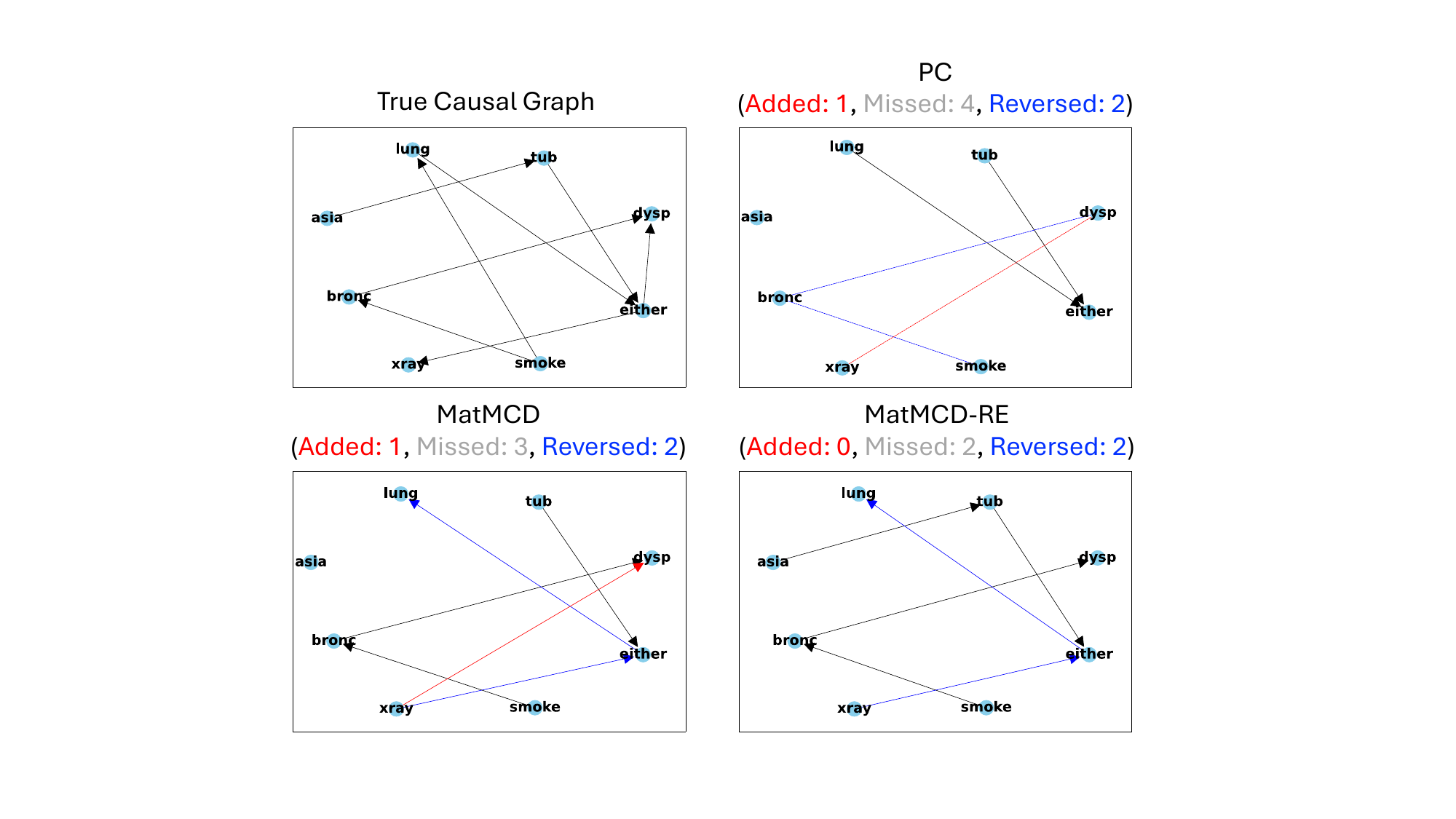}
\caption{Causal Graph visualization for \texttt{Asia} dataset.}\label{fig.vis4}
\end{figure}

\begin{figure}[!t]
\centering
\includegraphics[width=0.8\linewidth]{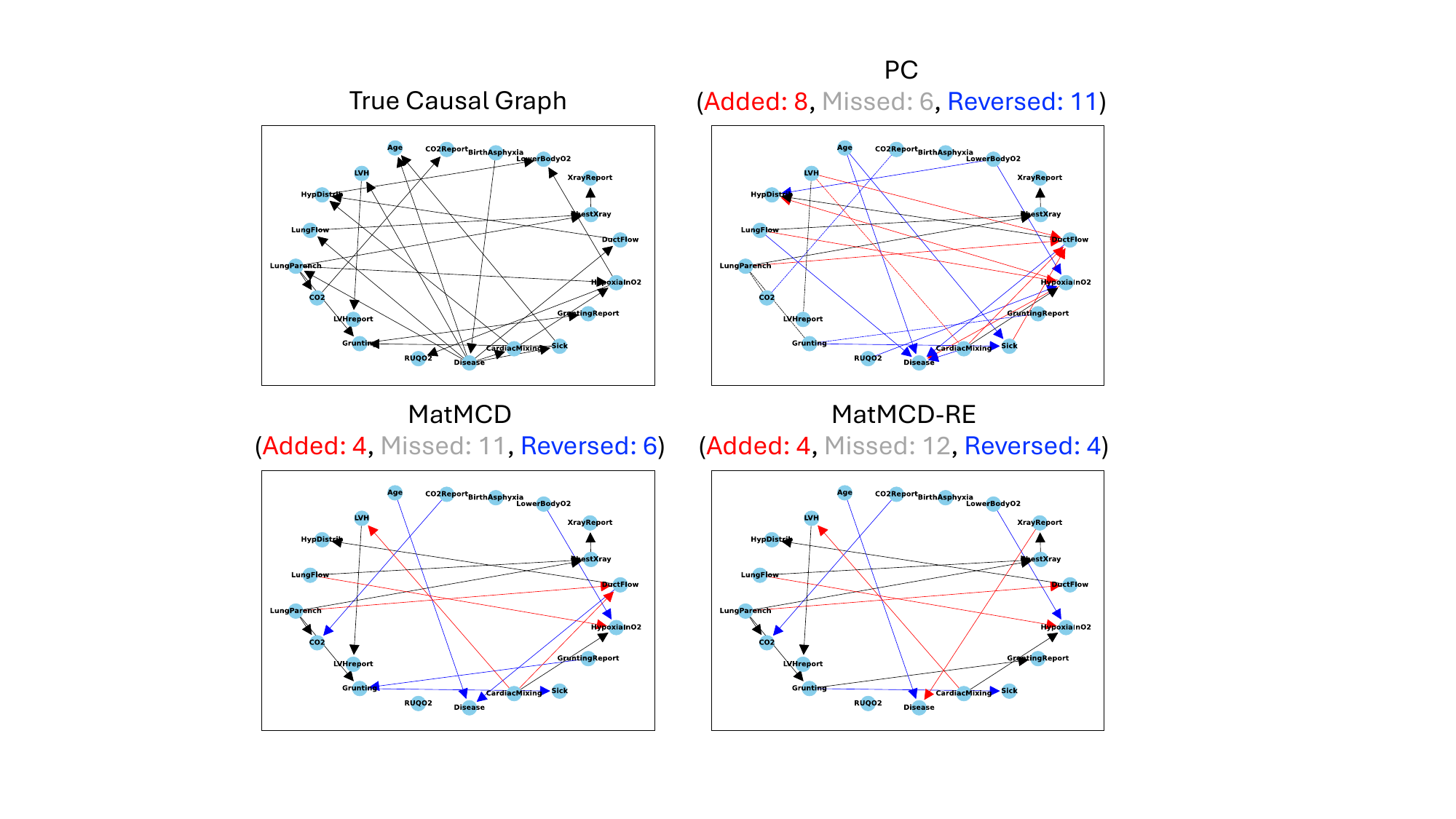}
\caption{Causal Graph visualization for \texttt{Child} dataset.}\label{fig.vis5}
\end{figure}

In this section, we present the visualization results of datasets \texttt{DWDClimate}, \texttt{SachsProtein}, \texttt{Asia} and \texttt{Child} in 
Fig. \ref{fig.vis2}, Fig. \ref{fig.vis3}, Fig. \ref{fig.vis4}, and Fig. \ref{fig.vis5}. Each figure includes the true causal graph, the causal graph predicted by the PC algorithm, the causal graph generated by \method, and the causal graph generated by \method-RE. In the figures, \textcolor{red}{red} arrows indicate edges that were added by error, \textcolor{blue}{blue} arrows denote edges that were reversed by error. In each figure, we also included the number of wrongly \textcolor{red}{Added} edges, \textcolor{gray}{Missed} edges, and \textcolor{blue}{Reversed} edges for comparison.

Compared to the PC algorithm, \method(-RE) yields causal graphs that resemble the true causal graph better, with in general less numbers of wrongly Added, Missed, and Reversed edges. This advantage stems from the LLM agents' ability to impose multi-modal data empowered knowledge-driven constraints for screening erroneous edges. In contrast, the PC algorithm only relies on correlations of the observational data between variables, rendering it more error-prone.

Furthermore, unlike the PC algorithm, which relies on an undirected graph structure to determine the direction of causal edges, the incorporation of LLM-provided knowledge offers precise guidance in establishing causal directions. Therefore, the proposed \method(-RE) exhibits fewer reversed edges on most of the datasets, particularly on the datasets with more accessible multi-modal data ({\em e.g.}, the non-biomedical datasets).


\clearpage

\section{Prompt Templates}

In this section, we provide the prompt templates for the Search LLM and Summary LLM in \modulea\ ($\S$\ref{sec.data_agent}), the Knowledge LLM and Constraint LLM in \moduleb\ ($\S$\ref{sec.cc_agent}). Additionally, we include the prompt template for the single-round search used in the Ablation Analysis in Table \ref{tab: Ablation experiment}(a) and the prompt template for the \moduleb\ without Knowledge LLM used in Ablation Analysis in Table \ref{tab: Ablation experiment}(b).

\subsection{Data Augmentation Agent (\modulea)}\label{app:DA_agent prompt}

\vspace{0.3cm}

\noindent
\fbox{
\centering
  \begin{minipage}{\dimexpr \linewidth-4\fboxsep-2\fboxrule}
    \begin{center}\textbf{Search LLM}\end{center}

    \vspace{0.2cm}
    
        \texttt{Assume you have no prior knowledge about the <\var{dataset name}> dataset and its nodes <\var{list of node names}>. You need to generate queries for others who can search for information that will help you summarize the dataset and the nodes, and help you recognize the relationships between the nodes.}

        \vspace{0.2cm}
        
        \texttt{Split your query into multiple sub-queries. Ensure each query is specific, clear, and easy to be used for search. If you have inquired about this topic before, your previous queries are:}

        \vspace{0.2cm}
        
        \texttt{<\var{list of previous queries}>}

        \vspace{0.2cm}

        \texttt{Please try to avoid repeating these queries. Generate a new query to get more information. Format your query as follows:}

        \vspace{0.2cm}
        
        \texttt{Search Query: <\var{new query}>}

        \vspace{0.2cm}
        
        \texttt{You should generate only one query at a time. 
        If you think no additional queries are needed, please state `No query needed'.}
    \end{minipage}
}

\vspace{0.3cm}

\noindent
\fbox{
\centering
  \begin{minipage}{\dimexpr \linewidth-4\fboxsep-2\fboxrule}
    \begin{center}\textbf{Summary LLM (with RAG)}\end{center}

    \vspace{0.2cm}
    
        \texttt{Please provide a summary of the <\var{dataset name}> dataset and its nodes including <\var{list of node names}> using the information from the RAG Database. 
        Your response should include detailed information on the dataset and each of its nodes.}

        \vspace{0.2cm}
        
        \texttt{Follow the structured format below in your answer:}

        \vspace{0.2cm}
        
        \texttt{- Dataset Summary: <a general summary of the dataset>}

        \vspace{0.2cm}
        
        \texttt{- Summary of <node name>: <a detailed summary of this node>}

        \vspace{0.2cm}
        
        \texttt{- Summary of <node name>: <a detailed summary of this node>}

        \vspace{0.2cm}
        
        \texttt{Additionally, try to include information on the relationships between the nodes after you have summarized each one. Make sure to cover all the nodes mentioned.}
    \end{minipage}
}

\clearpage

\noindent
\fbox{
  \begin{minipage}{\dimexpr \linewidth-4\fboxsep-2\fboxrule}
\begin{center}\textbf{Summary LLM (with Log)}\end{center}

\vspace{0.2cm}

\texttt{In <\var{dataset name}> dataset, the following entities are included: <\var{list of node names}>. The log information about the entity <\var{node name}> is provided in the following format:}

\vspace{0.2cm}

\texttt{<\var{log format template}>}

\vspace{0.2cm}

\texttt{Below are the information in the log following the above format:}

\vspace{0.2cm}

\texttt{<\var{The list of node's event templates and examples}>}

\vspace{0.2cm}

\texttt{Based on the above information, please provide your summary of this entity: <\var{node name}>, including:}

\vspace{0.2cm}

\texttt{- The role of the entity: <\var{node name}> in the system}

\vspace{0.2cm}

\texttt{- The key events that need to be noticed. Need to pay attention to both frequent and infrequent occurrences}

\vspace{0.2cm}

\texttt{- The status of this entity}

\vspace{0.2cm}

\texttt{- The relationships this entity: <\var{node name}> has with other entities in the above entity list. Only include the most likely ones.}

\vspace{0.2cm}

\texttt{You only need to provide the information about this entity: <\var{node name}>. Your response should be in the following format:}

\vspace{0.2cm}

\texttt{The name of the entity: <\var{node name}>}

\vspace{0.2cm}

\texttt{Role of the entity: <\var{role}>}

\vspace{0.2cm}

\texttt{Key events that need to be noticed: <\var{key events}>}

\vspace{0.2cm}

\texttt{The status of this entity: <\var{status}>}

\vspace{0.2cm}

\texttt{The relationships of <\var{node name}>: <\var{relationships}>}

\vspace{0.2cm}

\texttt{Your response:}

\vspace{0.2cm}

\end{minipage}
}

\subsection{Causal Constraint Agent (\moduleb)}\label{app:CC-Agent prompt}

\vspace{0.3cm}

\noindent
\fbox{
\begin{minipage}{\dimexpr \linewidth-4\fboxsep-2\fboxrule}
\begin{center}\textbf{Knowledge LLM}\end{center}

\vspace{0.2cm}

\texttt{We want to perform causal discovery on <\var{dataset name}> , the summary of dataset: <\var{dataset information}>. Considering <\var{list of node names}> as variables. We have conducted the statistical causal discovery with <\var{SCD algorithm name}> algorithm.}

\vspace{0.2cm}

\texttt{The edges and their coefficients of the causal structure suggested by the statistical discovery are as follows:}

\vspace{0.2cm}

\texttt{<\var{adjacency list of the causal graph}>}

\vspace{0.2cm}

\texttt{Based on the information above, it seems that changes in <\var{node name i}> have <\var{a/no}> direct impact on <\var{node name j}>. In addition, here is the information of <\var{node name i}> and <\var{node name j}> from reliable sources:}

\vspace{0.2cm}

\texttt{<\var{information about node i}>}

\vspace{0.2cm}

\texttt{<\var{information about node j}>}

\vspace{0.2cm}

\texttt{Your task is to interpret this result from a domain knowledge perspective and determine whether this statistically suggested hypothesis is plausible in the context of the domain. Please provide an explanation that leverages your expert knowledge on the causal relationship between <\var{node name i}> and <\var{node name j}>, and assess the correctness of this causal discovery result.}

\vspace{0.2cm}

\texttt{Your response should consider the relevant factors and provide a reasonable explanation based on your understanding of the domain.}
\end{minipage}
}



\vspace{0.3cm}


  

  

  

  

  

  

\noindent
\fbox{
  \begin{minipage}{\dimexpr \linewidth-4\fboxsep-2\fboxrule}
  \begin{center}\textbf{Constraint LLM}\end{center}

  \vspace{0.2cm}
  
  \texttt{Provide your <\var{K}> best guesses and the probability that each is correct (0.0 to 1.0) for the following question. Give ONLY the guesses and probabilities, no other words or explanation. Each guess should infer the relationship step by step and finally end with <Yes> or <No>. For example:}

  \vspace{0.2cm}
  
  \texttt{G1: <the first most likely guess, infer the relationship step by step and end with <Yes> or <No> >}

  \vspace{0.2cm}
  
  \texttt{P1: <the probability between 0.0 and 1.0 that G1 is correct, without any extra comments; just the probability!>}

  \vspace{0.2cm}
  
  \texttt{G2: <the second most likely guess, infer the relationship step by step and end with <Yes> or <No> >}

  \vspace{0.2cm}
  
  \texttt{P2: <the probability between 0.0 and 1.0 that G2 is correct, without any extra comments; just the probability!>}

  \vspace{0.2cm}

  \texttt{The question is: here is the explanation from an expert in the field of <\var{dataset name}> regarding the causal relationship between <\var{node name i}> and <\var{node name j}>:}

  \vspace{0.2cm}

  \texttt{<\var{domain knowledge output of Knowledge LLM}>}

  \vspace{0.2cm}

  \texttt{Considering the information above, if <\var{node name i}> is modified, will it have a direct impact on <\var{node name j}>?}
  \end{minipage}
}

\newpage
\subsection{Single-Round Search}\label{app:single round}

In the implementation of single-round search, 
SearchGPT\footnote{\url{https://github.com/Wilson-ZheLin/SearchGPT}} was used in combination with the prompt below to directly search for pertinent information about the dataset and nodes, thus the LLM is not required to generate a query.\\

\noindent
\fbox{
  \begin{minipage}{\dimexpr \linewidth-4\fboxsep-2\fboxrule}
\begin{center}\textbf{Single-Round Search}\end{center}

\vspace{0.2cm}

\texttt{Please provide detailed information of dataset <\var{dataset name}> and its variables, including <\var{list of node names}>. Try to search for more information if possible. You should provide information about each variable and their relationships with each other. Please provide detailed information.}

\vspace{0.2cm}

\texttt{Follow the structured format below in your answer:}

\vspace{0.2cm}

\texttt{The summary of dataset <\var{dataset name}>:}

\vspace{0.2cm}

\texttt{- Dataset Summary: <a general summary of the dataset>}

\vspace{0.2cm}

\texttt{- Summary of <node name>: <a detailed summary of this node>}

\vspace{0.2cm}

\texttt{- Summary of <node name>: <a detailed summary of this node>}

\vspace{0.2cm}

\texttt{Additionally, try to include information on the relationships between the nodes after you have summarized each one. Make sure to cover all the nodes mentioned.}

\end{minipage}
}

\vspace{1cm}

\subsection{\moduleb\ without Knowledge LLM}\label{app.ccagent}

\noindent
\fbox{
\begin{minipage}{\dimexpr \linewidth-4\fboxsep-2\fboxrule}
\begin{center}\textbf{\moduleb\ without Knowledge LLM}\end{center}

\vspace{0.2cm}

\texttt{We want to perform causal discovery on \textless\var{dataset name}\textgreater, the summary of dataset: \textless\var{dataset information}\textgreater. Considering \textless\var{list of node names}\textgreater\ as variables. We have conducted the statistical causal discovery with \textless\var{SCD algorithm name}\textgreater\ algorithm.}

\vspace{0.2cm}

\texttt{The edges and their coefficients of the causal structure suggested by the statistical discovery are as follows:}

\vspace{0.2cm}

\texttt{\textless\var{adjacency list of the causal graph}\textgreater}

\vspace{0.2cm}

\texttt{Based on the information above, it seems that changes in <\var{node name i}> have <\var{a/no}> direct impact on <\var{node name j}>. In addition, here is the information of <\var{node name i}> and <\var{node name j}> from reliable sources:}

\vspace{0.2cm}

\texttt{<\var{information about node i}>}

\vspace{0.2cm}

\texttt{<\var{information about node j}>}

\vspace{0.2cm}

\texttt{Do you think that a change in \textless\var{node name i}\textgreater\ will lead to a change in \textless\var{node name j}\textgreater, which means \textless\var{node name i}\textgreater\ has a causal influence on \textless\var{node name j}\textgreater?}

\vspace{0.2cm}

\texttt{Please answer this question with ⟨Yes⟩ or ⟨No⟩. No other responses are needed. Your response should be in the following format: ⟨Yes⟩ or ⟨No⟩.}

\end{minipage}
}


\newpage

\section{Examples of LLM Response}

\subsection{Data Augmentation Agent (\modulea)}\label{app:DA-Agent example}
In this section, we provide the response of LLMs during the search process of \modulea\ on the \texttt{AutoMPG} dataset.

\vspace{0.3cm}

\noindent
\fbox{
  \begin{minipage}{\dimexpr \linewidth-4\fboxsep-2\fboxrule}
        \begin{minipage}[t]{\textwidth}
        \textbf{Query 1 (from Search LLM)}: what is the auto mpg dataset, and what are the variables displacement, mpg, horsepower, weight, and acceleration?
        \end{minipage}
        \hrule\vspace{5pt}
        \begin{minipage}[t]{\textwidth}
        \textbf{Answer 1}: The auto mpg dataset is a well-known dataset in the field of machine learning and statistics, primarily used for regression analysis and predictive modeling. it contains information about various automobiles, specifically focusing on their fuel efficiency measured in miles per gallon (mpg). the dataset includes several variables that are crucial for understanding vehicle performance, including displacement, mpg, horsepower, weight, and acceleration.\\

        1. \textbf{displacement}: this variable refers to the total volume of all the cylinders in an engine, typically measured in cubic inches or liters. it is a key indicator of engine size and capacity. generally, a larger displacement can lead to more power output, but it may also result in lower fuel efficiency.
        
        2. \textbf{mpg (miles per gallon)}: this is the target variable in the dataset, representing the fuel efficiency of the vehicle. higher mpg values indicate better fuel efficiency, meaning the vehicle can travel further on a gallon of fuel. the relationship between mpg and other variables is often inverse; as engine size (displacement) and weight increase, mpg tends to decrease.
        
        3. \textbf{horsepower}: this variable measures the power output of the engine. higher horsepower typically correlates with better acceleration and overall performance. however, like displacement, higher horsepower can also lead to lower fuel efficiency, as more power often requires more fuel.
        
        4. \textbf{weight}: the weight of the vehicle is another critical factor affecting performance. heavier vehicles generally require more power to accelerate, which can negatively impact mpg. there is often a negative correlation between weight and mpg; as weight increases, fuel efficiency tends to decrease.
        
        5. \textbf{acceleration}: this variable measures how quickly a vehicle can increase its speed, usually defined as the time taken to go from 0 to 60 mph. Acceleration is influenced by both horsepower and weight; more powerful engines can accelerate faster, while heavier vehicles may struggle to achieve the same acceleration.\\
        
        In summary, the relationships among these variables can be summarized as follows:\\\\
        - \textbf{Displacement and Horsepower}: Generally, larger displacement engines produce more horsepower.\\
        - \textbf{Horsepower and Acceleration}: Higher horsepower typically results in better acceleration.\\
        - \textbf{Weight and MPG}: Increased weight usually leads to lower MPG.\\
        - \textbf{Displacement and MPG}: Larger engines (higher displacement) often result in lower MPG due to increased fuel consumption.\\
        
        Understanding these relationships is crucial for analyzing vehicle performance and making informed decisions regarding vehicle design and efficiency improvements.
        \end{minipage}
    \end{minipage}
}\newpage

\noindent
\fbox{
  \begin{minipage}{\dimexpr \linewidth-4\fboxsep-2\fboxrule}
        \begin{minipage}[t]{\textwidth}
        \textbf{Query 2 (from Search LLM)}: What are the relationships between its variables such as Displacement, MPG, Horsepower, Weight, and Acceleration?           
        \end{minipage}
        \hrule\vspace{5pt}
        \begin{minipage}[t]{\textwidth}
        \textbf{Answer 2}: The Auto MPG dataset is a well-known dataset that contains information about various attributes of automobiles and their fuel consumption, specifically measured in miles per gallon (MPG). The dataset includes five key continuous variables: \textbf{Weight}, \textbf{Displacement}, \textbf{Horsepower}, \textbf{Acceleration}, and \textbf{MPG}. It is often used in data analysis and machine learning to understand the factors that influence fuel efficiency in cars.\\
        
        \textbf{Relationships Between Variables}\\
        1. \textbf{Displacement}: Displacement refers to the engine size of the car, and it has been found to have a strong negative correlation with MPG. This means that as the displacement increases, the MPG tends to decrease. This relationship is intuitive, as larger engines typically consume more fuel.
        
        2. \textbf{Horsepower}: Similar to displacement, horsepower also shows a negative causal effect on MPG. Higher horsepower engines are generally less fuel-efficient, which is supported by statistical analysis indicating a significant negative relationship between horsepower and MPG.
        
        3. \textbf{Weight}: The weight of the car is another critical factor affecting fuel efficiency. Heavier cars tend to have lower MPG values, as more energy is required to move a heavier vehicle. This relationship is also confirmed through causal inference results, indicating that increasing the weight of a car significantly reduces its fuel efficiency.
        
        4. \textbf{Acceleration}: Acceleration has a complex relationship with MPG. While faster acceleration can be associated with lower fuel efficiency, the analysis suggests that the relationship is not as straightforward as with displacement and horsepower. The causal effect of acceleration on MPG is negative, indicating that cars designed for quicker acceleration may sacrifice fuel efficiency.
        
        5. \textbf{MPG}: MPG is a key indicator for measuring vehicle fuel efficiency, reflecting the distance a vehicle can travel on a certain amount of fuel. This indicator is influenced by various factors, including weight, displacement, horsepower, and acceleration. \\
        
        \textbf{Summary of Findings}
        
        - \textbf{Negative Correlations}: Displacement, horsepower, and weight are negatively correlated with MPG, meaning that increases in these variables lead to decreases in fuel efficiency.\\
        - \textbf{Complex Relationships}: Acceleration has a negative effect on MPG, but the relationship is influenced by other factors and may not be as direct as with weight or displacement.\\
        
        In conclusion, the Auto MPG dataset provides valuable insights into how various automobile attributes affect fuel consumption. Understanding these relationships can help in designing more fuel-efficient vehicles and improving overall automotive performance.
        \end{minipage}
    \end{minipage}
}\newpage

\noindent
\fbox{
  \begin{minipage}{\dimexpr \linewidth-4\fboxsep-2\fboxrule}
        \begin{minipage}[t]{\textwidth}
        \begin{center}\textbf{Example Summary of \texttt{AutoMPG} Dataset}\end{center}

        \vspace{0.2cm}
        
        \textbf{Dataset Summary}: The Auto MPG dataset is a well-known dataset in the field of machine learning and statistics, primarily used for regression analysis and predictive modeling. It contains information about various automobiles, specifically focusing on their fuel efficiency measured in miles per gallon (MPG). The dataset includes several variables that are crucial for understanding vehicle performance, including Displacement, Mpg, Horsepower, Weight, and Acceleration.

        \vspace{0.2cm}
        
        \texttt{/* Each of the following summaries describes a variable and its relationships with other variables */}

        - - -

        \vspace{0.2cm}
        
        \textbf{Summary of Displacement}: Displacement refers to the total volume of all the cylinders in an engine, typically measured in cubic inches or liters. It is a key indicator of engine size and capacity. Generally, a larger displacement can lead to more power output, but it may also result in lower fuel efficiency.
        
        - - -

        \vspace{0.2cm}
        
        \textbf{Summary of Mpg}: Mpg (Miles per Gallon) is the target variable in the dataset, representing the fuel efficiency of the vehicle. Higher MPG values indicate better fuel efficiency, meaning the vehicle can travel further on a gallon of fuel. The relationship between MPG and other variables is often inverse; as engine size (Displacement) and weight increase, MPG tends to decrease.
        
        - - -

        \vspace{0.2cm}
        
        \textbf{Summary of Horsepower}: Horsepower measures the power output of the engine. Higher horsepower typically correlates with better acceleration and overall performance. However, like displacement, higher horsepower can also lead to lower fuel efficiency, as more power often requires more fuel.
        
        - - -

        \vspace{0.2cm}
        
        \textbf{Summary of Weight}: The weight of the vehicle is another critical factor affecting performance. Heavier vehicles generally require more power to accelerate, which can negatively impact MPG. There is often a negative correlation between weight and MPG; as weight increases, fuel efficiency tends to decrease.
        
        - - -

        \vspace{0.2cm}
        
       \textbf{Summary of Acceleration}: Acceleration measures how quickly a vehicle can increase its speed. It is influenced by both horsepower and weight; more powerful engines can accelerate faster, while heavier vehicles may struggle to achieve the same acceleration.
       \end{minipage}
    \end{minipage}
}

\newpage

\subsection{Causal Constraint Agent (\moduleb)}\label{app:CC-Agent example}
In this section, we provide two examples of knowledge generated by the Knowledge LLM in \moduleb, including cases where the Knowledge LLM agrees with and opposes the results of the SCD algorithm.\\\\

\noindent
\fbox{
  \begin{minipage}{\dimexpr \linewidth-4\fboxsep-2\fboxrule}
        \begin{minipage}[t]{\textwidth}
        \textbf{Domain Knowledge from Displacement $\rightarrow$ MPG}
        \end{minipage}
        \hrule\vspace{5pt}
        \begin{minipage}[t]{\textwidth}
        The findings from the statistical causal discovery using the PC algorithm suggest a complex interplay between the variables in the AutoMPG dataset, particularly highlighting the relationships between Displacement, Horsepower, Weight, Acceleration, and Mpg. To interpret these results from a domain knowledge perspective, let's break down the relationships and their implications for gasoline consumption.\\\\
        \textbf{Understanding the Variables}\\
        - \textbf{Displacement}: This is a measure of the engine's size and capacity. Generally, larger displacement engines can produce more power, which can be beneficial for performance. However, larger engines often consume more fuel, leading to lower fuel efficiency (MPG).\\
        - \textbf{Mpg (Miles per Gallon)}: This is the key measure of fuel efficiency. Higher MPG values indicate that a vehicle can travel further on a gallon of fuel, which is desirable for cost savings and environmental considerations.\\\\
        \textbf{Interpretation of the Findings}

        The conclusion that changes in Displacement do not directly affect Mpg is plausible and aligns with our understanding of automotive engineering and fuel efficiency. Here are some key considerations:

        1. \textbf{Indirect Relationships}: While Displacement itself does not directly influence Mpg, it affects other variables (Horsepower, Weight, Acceleration) that subsequently impact MPG. This indicates a multi-layered causal structure where Displacement plays a foundational role, but its effects are mediated through other factors.

        2. \textbf{Complex Interactions}: The relationships among these variables are not straightforward. For example, a vehicle with a larger engine might be heavier and more powerful, but if it is designed with fuel efficiency in mind (e.g., hybrid technology), it could achieve a higher MPG despite the larger displacement.
        
        3. \textbf{Empirical Evidence}: Empirical studies in automotive engineering support the notion that while larger engines tend to be less fuel-efficient, the design choices and technologies implemented in vehicles can significantly alter this relationship. For instance, advancements in engine technology (like turbocharging) can allow for smaller engines to produce more power while maintaining or improving fuel efficiency.\\\\
        \textbf{Conclusion}\\
        The causal discovery results are consistent with domain knowledge. Displacement does not directly affect MPG, but it influences other factors that collectively determine fuel efficiency. This nuanced understanding reflects the complexity of automotive design and performance, where multiple variables interact in ways that may not be immediately apparent. 
        
        Overall, the findings emphasize the importance of considering the entire system of vehicle characteristics rather than isolating single variables when assessing gasoline consumption and fuel efficiency. This insight is crucial for developing more efficient vehicles and informing consumer choices.
        \end{minipage}
    \end{minipage}
}\newpage

\noindent
\fbox{
  \begin{minipage}{\dimexpr \linewidth-4\fboxsep-2\fboxrule}
        \begin{minipage}[t]{\textwidth}
        \textbf{Domain Knowledge from MPG $\rightarrow$ Weight}
        \end{minipage}
        \hrule\vspace{5pt}
        \begin{minipage}[t]{\textwidth}
        \textbf{Causal Relationships Overview}\\
        In analyzing causal relationships within the Auto MPG dataset, we expect to see significant impacts from vehicle characteristics like engine displacement, horsepower, and weight on MPG. However, the idea that MPG could directly influence a vehicle’s weight is counterintuitive based on principles of vehicle design and physics.\\\\
        \textbf{Assessment of the Claim ``MPG Influences Weight''}\\
        1.\textbf{Physical and Engineering Principles}: In conventional vehicle engineering, weight is primarily determined by the vehicle’s physical construction, such as body materials, engine size, and overall design. MPG is an outcome metric that reflects the fuel efficiency of a vehicle under specific conditions, not a factor that could in turn influence these physical characteristics.\\
        2.\textbf{Design Prioritization}: Vehicle designers prioritize meeting performance and safety standards before optimizing for fuel efficiency. This means that decisions about weight are made early in the design and manufacturing process, and MPG is a derivative outcome after these factors are considered. Thus, it is logically inconsistent to view MPG as a direct influencer of weight.\\
        3.\textbf{Difference Between Statistical Correlation and Causation}: Even if statistical analysis shows a correlation between MPG and weight, this relationship should not be misconstrued as causative. Improvements in MPG are often achieved through the use of lighter materials and more efficient engines, but these improvements primarily affect the design and manufacturing process, not MPG directly acting on weight.\\\\
        \textbf{Reevaluating the Role of MPG}\\
        \textbf{Indirect Influence}: Although there is an association between MPG and a vehicle’s weight, such as lighter vehicles often having higher MPG, this relationship is more about weight influencing MPG rather than MPG influencing weight. The use of lightweight materials and efficient technologies is primarily aimed at achieving better performance and environmental standards, which may indirectly improve MPG.\\\\
        \textbf{Conclusion}\\
        Based on fundamental principles of vehicle manufacturing and design, MPG does not directly influence a vehicle’s weight. MPG results from the interaction of various factors, not as a cause of changes in those factors. In analyzing automotive data, it is crucial to distinguish between outcomes and causes and understand how these variables interact in real engineering and design contexts. Therefore, suggesting that MPG influences vehicle weight is a misconception, reflecting a misunderstanding of the vehicle design process and causal relationships.
        \end{minipage}
    \end{minipage}
}\newpage

\end{document}